%% file: main.tex
\documentclass[twoside,11pt]{article}

\usepackage[abbrvbib, preprint]{jmlr2e}

\usepackage{amsmath,amsfonts}
\usepackage{algorithmic}
\usepackage{textcomp}
\usepackage{xcolor}

\usepackage{caption}
\usepackage{subcaption}

\usepackage{import}

\usepackage{comment}

\usepackage{bm}
\usepackage{mathtools}

\usepackage{adjustbox}

\newcommand*\rot{\rotatebox{90}}

\usepackage{booktabs}

\usepackage{multirow} 
\usepackage{soul}
\usepackage{changepage,threeparttable} 
\usepackage{array, makecell}

\usepackage{xargs}                      
\usepackage[colorinlistoftodos,prependcaption,textsize=tiny]{todonotes}

\newcommand{\nalgorithms}{$32$~}
\newcommand{\npyodalgorithms}{$27$~}
\newcommand{\ndatasets}{$52$~}
\newcommand{\nglobaldatasets}{$37$~}
\newcommand{\nlocaldatasets}{$12$~}

\newcommand{\imamdavenportoverall}{$18.076$~}

\newcommand{\imamdavenportcriticaloverall}{$0.620$~}

\newcommandx{\unsure}[2][1=]{\todo[linecolor=red,backgroundcolor=red!25,bordercolor=red,#1]{#2}}
\newcommandx{\change}[2][1=]{\todo[linecolor=blue,backgroundcolor=blue!25,bordercolor=blue,#1]{#2}}
\newcommandx{\addition}[2][1=]{\todo[linecolor=green,backgroundcolor=green!25,bordercolor=green,#1]{#2}}
\newcommandx{\thiswillnotshow}[2][1=]{\todo[disable,#1]{#2}}
\usepackage{lastpage}
\jmlrheading{23}{2023}{1-\pageref{LastPage}}{?/??; Revised ?/??}{?/??}{21-0000}{Bouman, Bukhsh, and Heskes}

\ShortHeadings{Unsupervised anomaly detection algorithms on real-world data: how many do we need?}{Bouman, Bukhsh, and Heskes}
\firstpageno{1}

\begin{document}

\title{Unsupervised anomaly detection algorithms on real-world data: how many do we need?}

\author{\name Roel Bouman \email roel.bouman@ru.nl\\
       \addr Institute for Computing and Information Sciences \\
       Radboud University \\
       Toernooiveld 212, 6525 EC Nijmegen, The Netherlands
       \AND
       \name Zaharah Bukhsh \email z.bukhsh@tue.nl\\
       \addr Industrial Engineering \& Innovation Sciences, Information Systems \\
       Eindhoven University of Technology \\
       Groene Loper 3, 5612 AE Eindhoven, The Netherlands   
       \AND
       \name Tom Heskes \email tom.heskes@ru.nl\\
       \addr Institute for Computing and Information Sciences \\
       Radboud University \\
       Toernooiveld 212, 6525 EC Nijmegen, The Netherlands
       }

\editor{??}

\maketitle

\begin{abstract}
In this study we evaluate \nalgorithms unsupervised anomaly detection algorithms on \ndatasets real-world multivariate tabular datasets, performing the largest comparison of unsupervised anomaly detection algorithms to date. On this collection of datasets, the $k$-thNN (distance to the $k$-nearest neighbor) algorithm significantly outperforms the most other algorithms. Visualizing and then clustering the relative performance of the considered algorithms on all datasets, we identify two clear clusters: one with ``local'' datasets, and another with ``global'' datasets. ``Local'' anomalies occupy a region with low density when compared to nearby samples, while ``global'' occupy an overall low density region in the feature space. On the local datasets the $k$NN ($k$-nearest neighbor) algorithm comes out on top. On the global datasets, the EIF (extended isolation forest) algorithm performs the best. Also taking into consideration the algorithms' computational complexity, a toolbox with these three unsupervised anomaly detection algorithms suffices for finding anomalies in this representative collection of multivariate datasets. By providing access to code and datasets, our study can be easily reproduced and extended with more algorithms and/or datasets. 
\end{abstract}


\section{Introduction}
Anomaly detection is the study of finding data points that do not fit the expected structure of the data. Anomalies can be caused by unexpected processes generating the data. In chemistry an anomaly might be caused by an incorrectly performed experiment, in medicine a certain disease might induce rare symptoms, and in predictive maintenance an anomaly can be indicative of early system failure. Depending on the application domain, anomalies have different properties, and may also be called by different names. Within the domain of machine learning (and hence also in this paper), anomaly detection is often used interchangeably with outlier detection. 

Unsupervised, data-driven, detection of anomalies is a standard technique in machine learning. Throughout the years, many methods, or algorithms, have been developed in order to detect anomalies. Some of these algorithms aim to solve specific issues, such as high dimensionality. Other methods try to detect anomalies in the general sense, and focus on high performance or low computational or memory complexity. Due to the many algorithms available, it is hard to determine which algorithm is best suited for a particular use case, especially for a user who is not intimately familiar with the field of anomaly detection.

Several studies have been performed to provide guidelines on when to apply which algorithm. Some review studies~\citep{malik2014comparative, ruff2021unifying}, give advice based on the theoretical properties of the algorithms. In recent years, several studies have been conducted that empirically compare a number of anomaly detection algorithms on a range of datasets. 

\citet{emmott2015meta} study 8 well-known algorithms on 19 datasets. They find Isolation Forest to perform the best overall, but recommend using ABOD (Angle-Based anomaly Detection) or LOF (Local anomaly Factor) when there are multiple clusters present in the data. 

\citet{campos2016evaluation} compare 12 $k$-nearest neighbours based algorithms, on 11 base datasets. They find LOF to significantly outperform a number of other methods, while KDEOS (Kernel Density Estimation anomaly Score) performs significantly worse than most algorithms. 

\citet{goldstein2016comparative} compare 19 algorithms on 10 datasets. Unlike \citeauthor{campos2016evaluation}, \citeauthor{goldstein2016comparative} perform no explicit optimization or selection, but rather evaluate the average performance over a range of sensible hyperparameter settings. With methods based on $k$-nearest neighbours generally giving stable results, \citeauthor{goldstein2016comparative} recommend $k$NN ($k$-nearest neighbours) for global anomalies, LOF for local anomalies, and HBOS (Histogram-Based anomaly Selection) in general (see ~\ref{subsection:types_of_outliers} for an explanation of global and local anomalies). \citeauthor{goldstein2016comparative} compare on a dataset basis, without any overall statistical analysis.

More recently, \citet{domingues2018comparative}, apply 14 algorithms on 15 datasets, some of which are categorical. They find IF (Isolation Forest) and robust KDE (Kernel Density Estimation) to perform best, but note that robust KDE is often too expensive too calculate for larger datasets. 

\citet{steinbuss2021benchmarking} propose a novel strategy for synthesizing anomalies in real-world datasets using several statistical distributions as a sampling basis. They compare 4 algorithms across multiple datasets derived from 19 base datasets, both using the original and synthesized anomalies. They find $k$NN and IF to work best for detecting global anomalies, and LOF to work best for local and dependency anomalies.
In the same year, \citet{soenen2021effect} study the effect of hyperparameter optimization strategies on the evaluation and propose to optimize hyperparameters on a small validation set, with evaluation on a much larger test set.
In their comparison of 6 algorithms on 16 datasets, IF performs the best, closely followed by CBLOF/u-CBLOF ((unweighted-)Cluster-Based Local Outlier Factor) and $k$NN, while OCSVM  (One-Class Support Vector Machine) performs worst unless optimized using a substantially larger validation set than the other algorithms. 

\citet{han2022adbench} performed an extensive comparison of anomaly detection methods, including supervised and semi-supervised algorithms. They compare 14 unsupervised algorithms on 47 tabular datasets using out-of-the-box hyperparameter settings. They subsample larger datasets to a maximum of 10.000 samples, duplicate samples for those datasets smaller than 1000 samples. They find no significant differences between unsupervised algorithms. While real-world datasets are being used, the anomalies in each dataset are generated synthetically according to 4 different type definitions (see section \ref{subsection:types_of_outliers}), and they compare the performance for each different type. Additionally, they have analyzed more complex benchmark datasets used in CV and NLP, such as CIFAR10~\citep{krizhevsky2009learning} and the Amazon dataset~\citep{he2016ups} by performing neural-based feature extraction.

Other studies are of a more limited scope, and cover for example methods for high-dimensional data~\citep{xu2018comparison}, or consider only ensemble methods~\citep{zimek2014ensembles}.

The studies done by \citet{campos2016evaluation, goldstein2016comparative,domingues2018comparative,steinbuss2021benchmarking,soenen2021effect,han2022adbench} have several limitations when used as a benchmark. Firstly, with the exception of \citeauthor{han2022adbench}, all studies were done on a rather small collection of datasets. Secondly, these studies cover only a small number of methods. \citeauthor{campos2016evaluation} compare only $k$NN-based approaches, while \citeauthor{goldstein2016comparative} fail to cover many of the methods that have gained traction in the last few years, such as IF~\citep{liu2008isolation} and variants thereof~\citep{hariri2019extended}. \citeauthor{soenen2021effect} consider just 6 commonly used methods, \citeauthor{steinbuss2021benchmarking} cover 4 methods and \citeauthor{han2022adbench} cover 14 unsupervised methods.

Some of these studies consider the performance on datasets containing specific types of anomalies, such as global or local anomalies. Specifically, \citeauthor{steinbuss2021benchmarking} look at the performance of different algorithms on datasets containing synthesized global, local, and dependency anomalies. Similarly, \citeauthor{han2022adbench} synthesize these three types of anomalies as well as cluster anomalies for use in their comparison. \citeauthor{goldstein2016comparative}'s study is, to the best of our knowledge, the only one that analyzes real-world, i.e., non-synthesized global and local anomalies. In particular, they analyze the `pen-local' and `pen-global' dataset, two variants of the same dataset where different classes were selected to obtain local and global anomalies specifically.

In practice, very little is known regarding what types of anomalies are present in commonly used benchmark datasets, and thus large scale comparisons on real-world data for specifics types are still missing.
In this study we apply a large number of commonly used anomaly detection methods on a large collection of multivariate datasets, to discover guidelines on when to apply which algorithms. We explicitly choose to perform no optimization of hyperparameters, so as to evaluate the performance of algorithms in a truly unsupervised manner. Instead, we evaluate every algorithm over a range of sensible hyperparameters, and compare average performances. This contrasts with \citet{soenen2021effect}, who perform extensive optimization on a small validation set and thereby supply guidelines for semi-supervised detection or active learning. Our approach rather is similar to that used by \citet{domingues2018comparative}, who also compare off-the-shelf performance. To the best of our knowledge, ours is the largest study of its kind performed so far.

\section{Background}
\subsection{Types of anomalies}
\label{subsection:types_of_outliers}
Many different definitions of anomalies and their properties exist, many of these have been defined in an isolated context, not considering the relationships with other definitions or properties~\citep{breunig2000lof}. More recently, a review by \citet{foorthuis2021nature} tried to unify definitions across multiple subdomains of anomaly detection in order to encompass all types of anomalies. These definitions however do not encompass many properties or types of anomalies, such as clustered or dependency anomalies. 

Rather than aiming to redefine every distinct type of anomaly, we treat anomalies as being able to have multiple, sometimes non-exclusive properties. Instead, we define four scales of non-exclusive properties which, when combined, encompass all types of anomalies found in multivariate tabular data in literature known to us.

\begin{figure}
     \centering
     \begin{subfigure}[b]{0.45\textwidth}
         \centering
         \includegraphics[width=\textwidth]{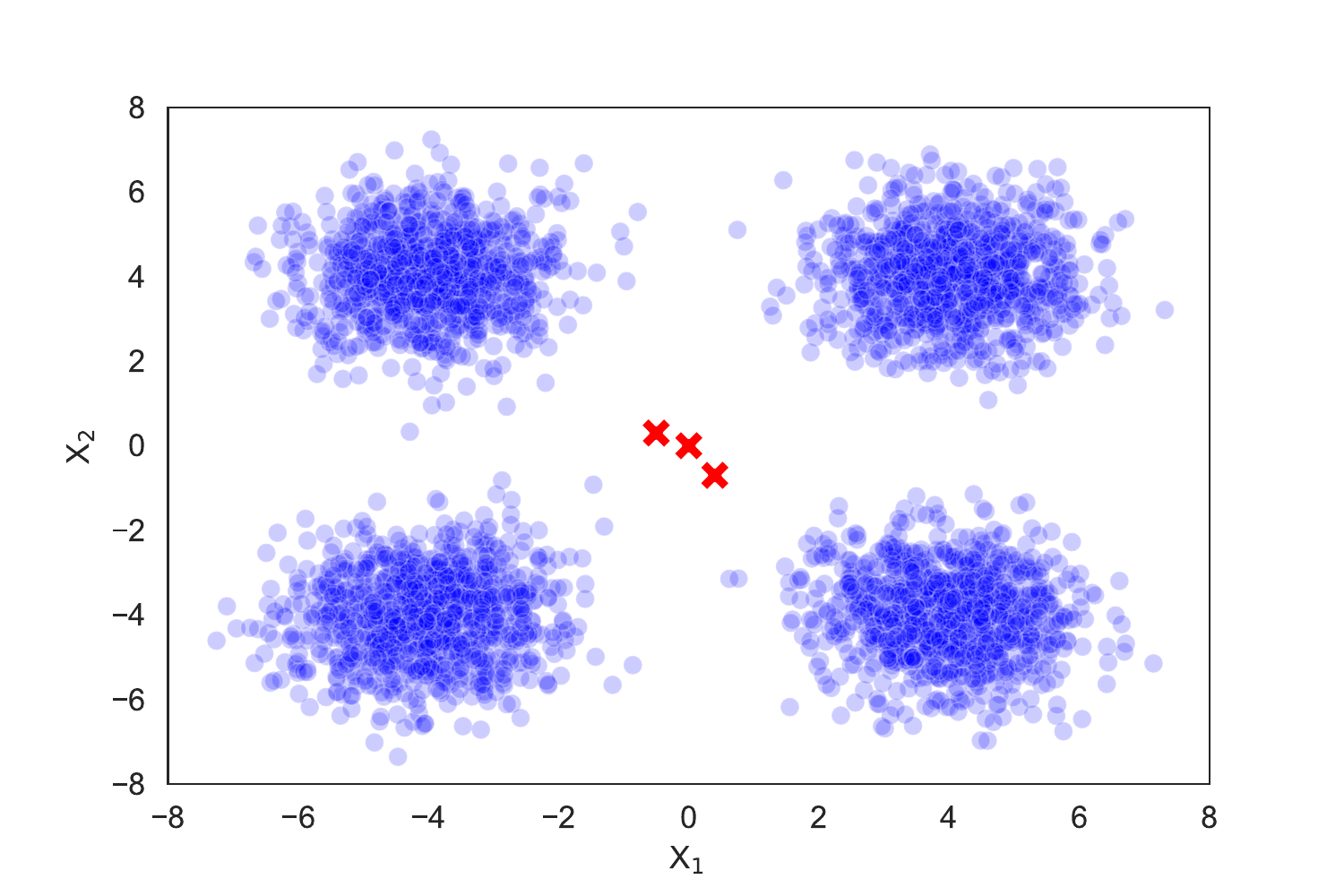}
         \caption{Enclosed anomaly example}
         \label{fig:enclosed_anomalies}
     \end{subfigure}
     \hfill
     \begin{subfigure}[b]{0.45\textwidth}
         \centering
         \includegraphics[width=\textwidth]{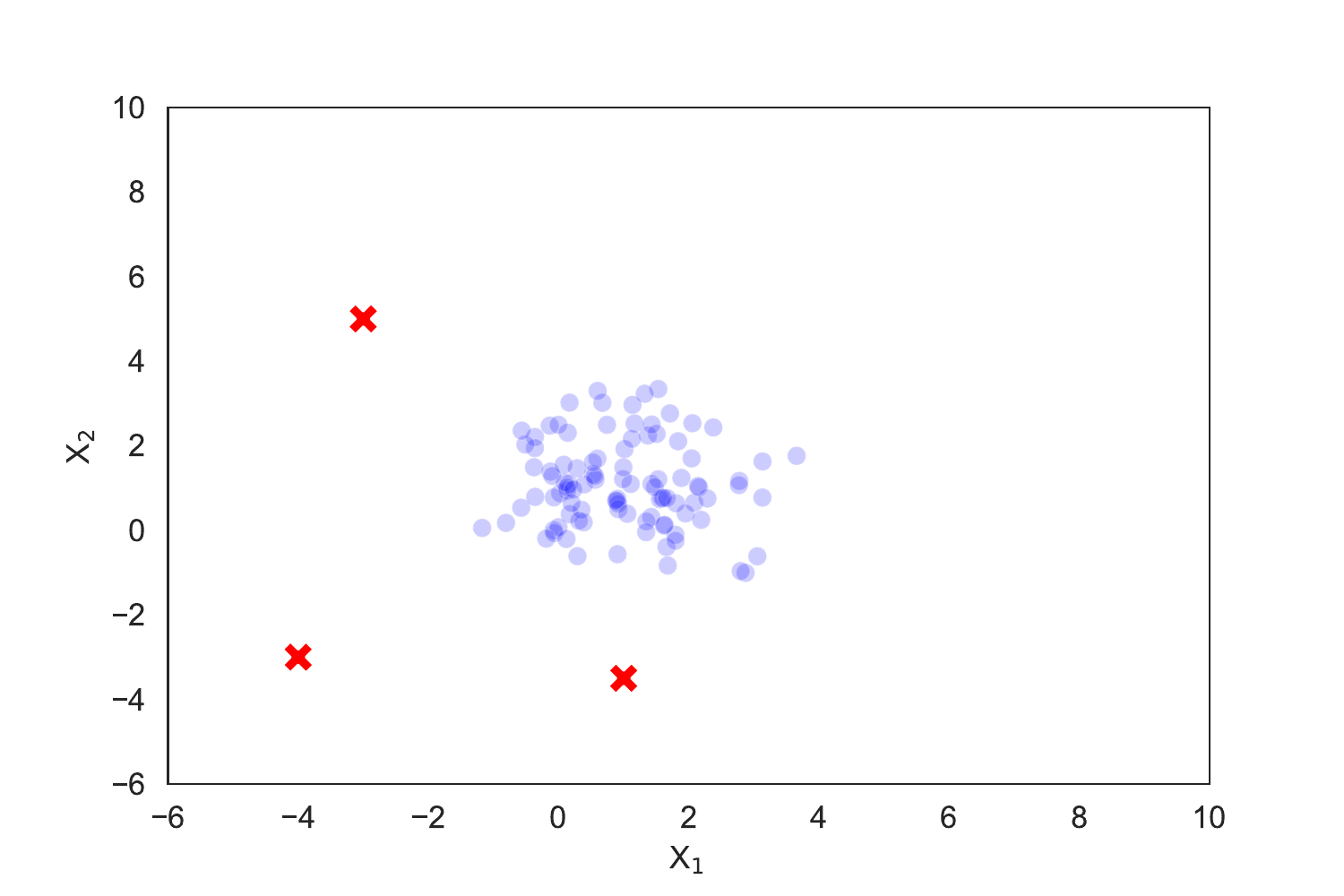}
         \caption{Peripheral anomaly example}
         \label{fig:peripheral_anomalies}
     \end{subfigure}
     \vfill
     \begin{subfigure}[b]{0.45\textwidth}
         \centering
         \includegraphics[width=\textwidth]{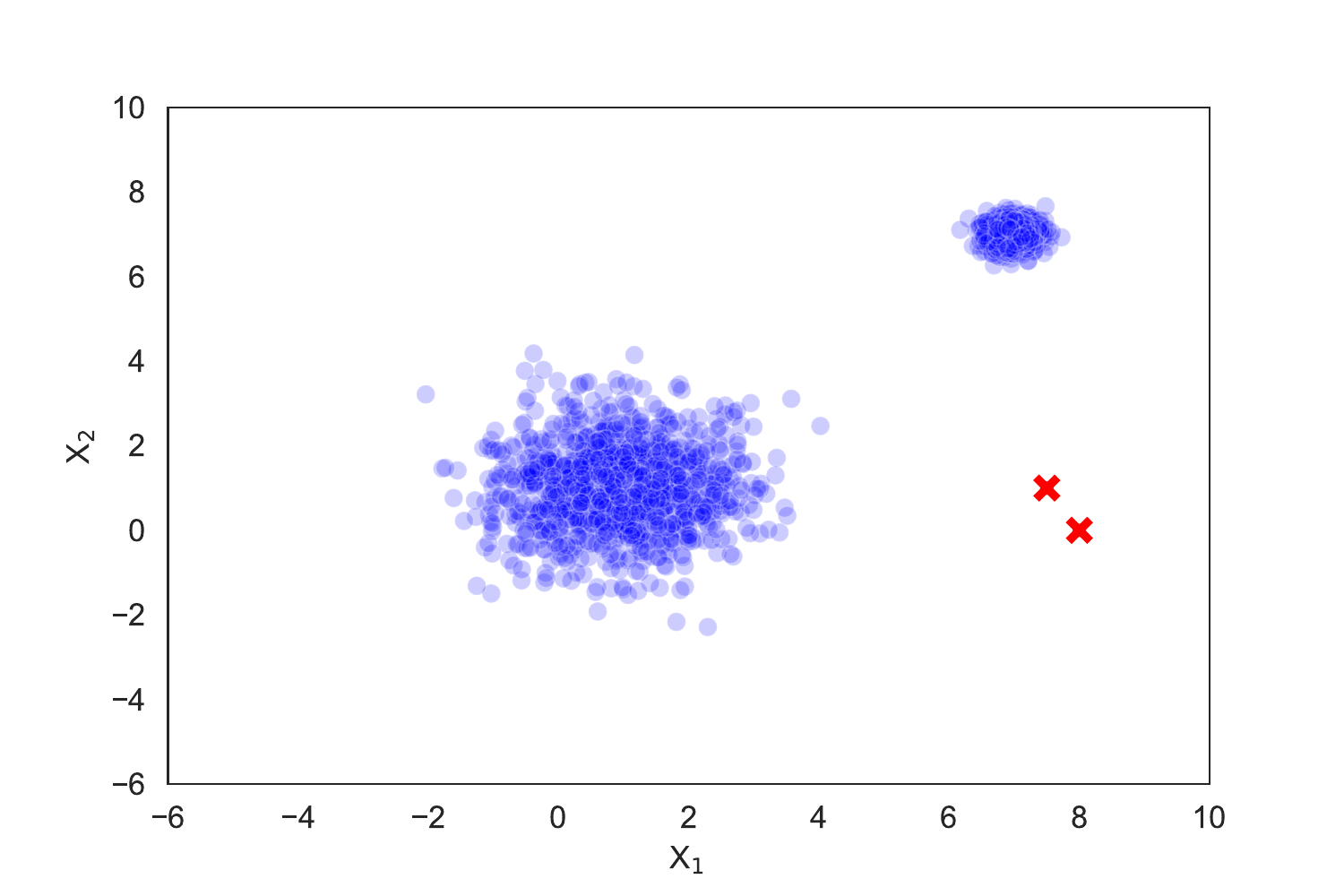}
         \caption{Global anomaly example}
         \label{fig:global_anomalies}
     \end{subfigure}
     \hfill
     \begin{subfigure}[b]{0.45\textwidth}
         \centering
         \includegraphics[width=\textwidth]{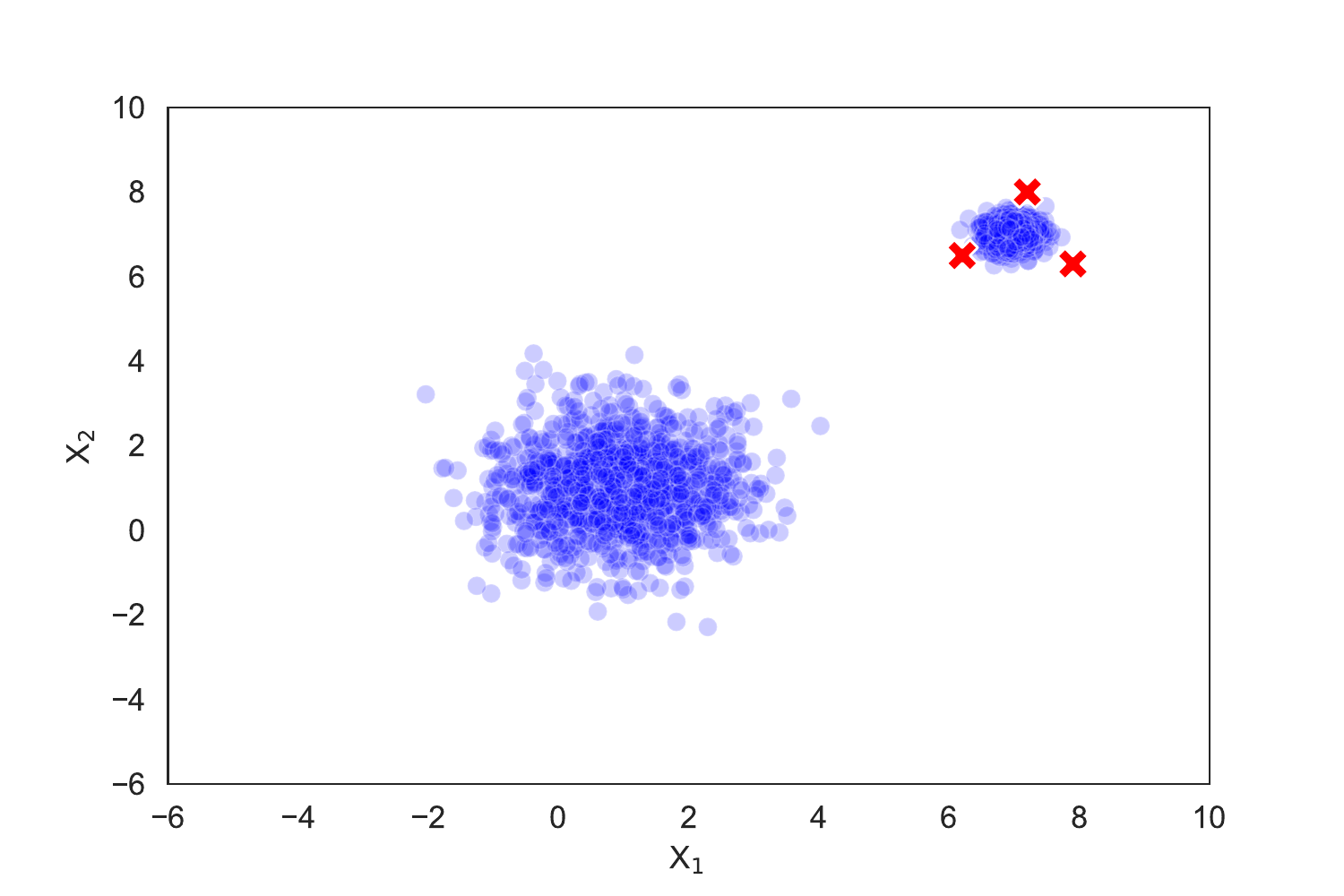}
         \caption{Local anomaly example}
         \label{fig:local_anomalies}
     \end{subfigure}
      \begin{subfigure}[b]{0.45\textwidth}
         \centering
         \includegraphics[width=\textwidth]{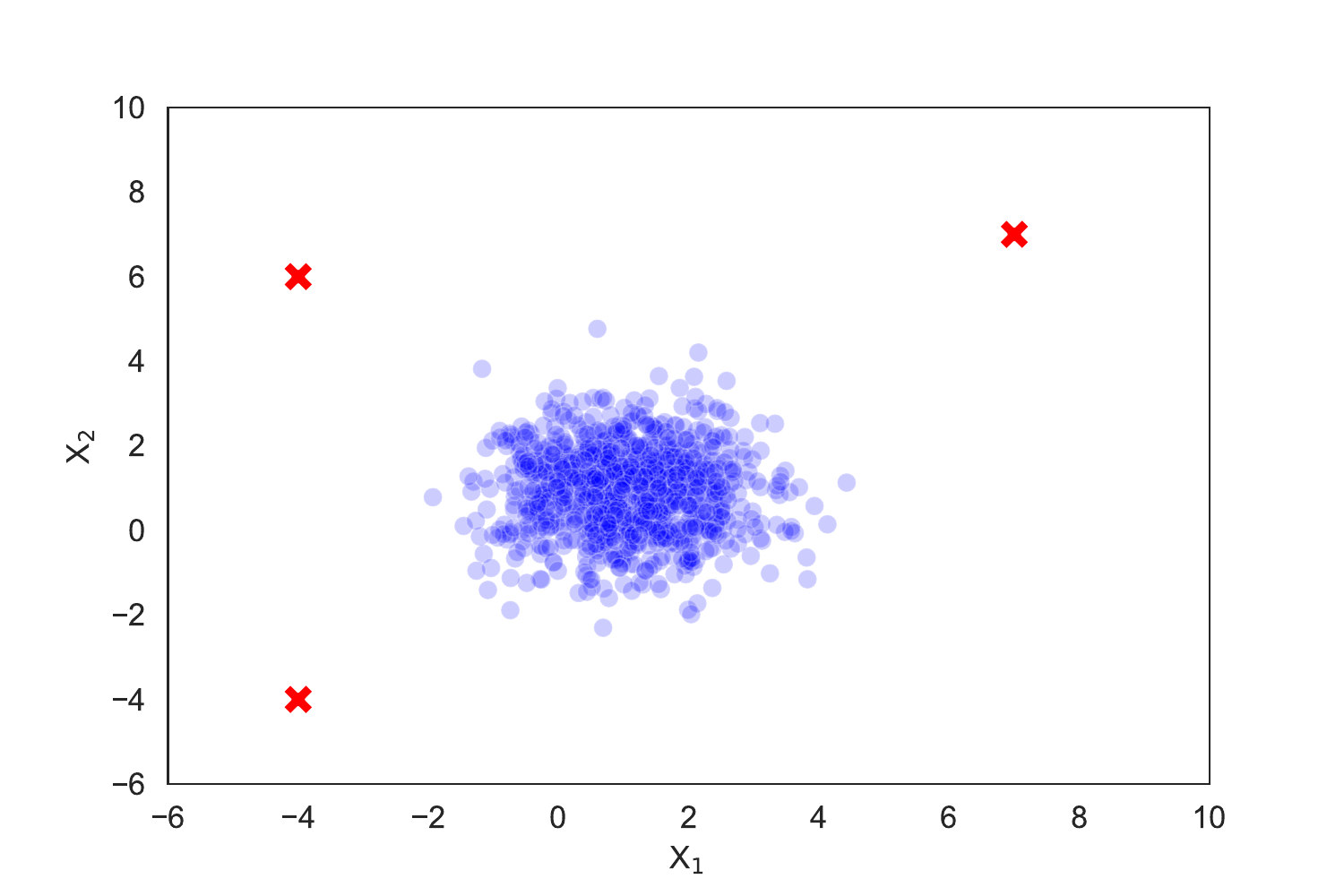}
         \caption{Isolated anomaly example}
         \label{fig:isolated_anomalies}
     \end{subfigure}
     \hfill
     \begin{subfigure}[b]{0.45\textwidth}
         \centering
         \includegraphics[width=\textwidth]{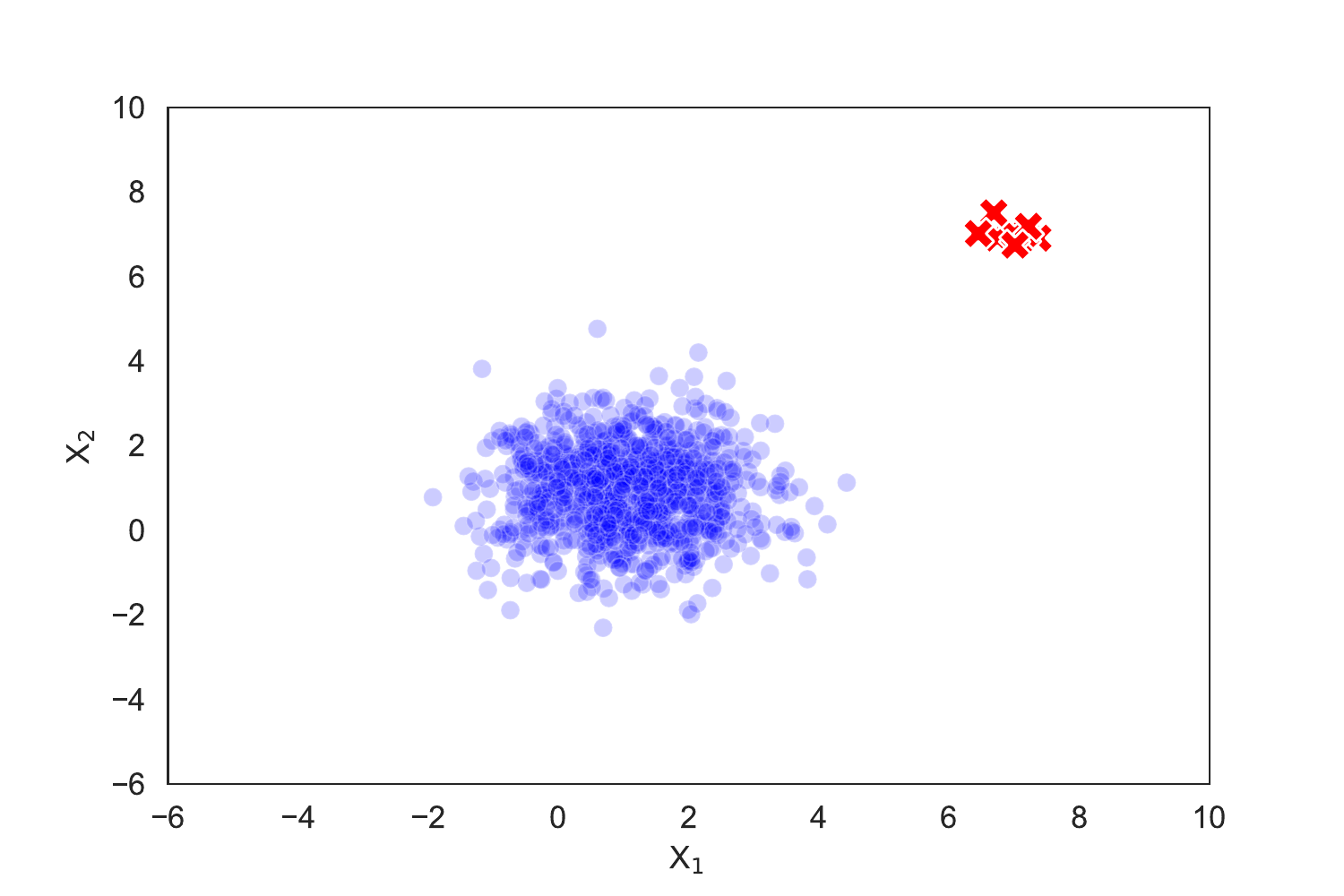}
         \caption{Clustered anomaly example}
         \label{fig:clustered_anomalies}
     \end{subfigure}
     \vfill
     \begin{subfigure}[b]{0.45\textwidth}
         \centering
         \includegraphics[width=\textwidth]{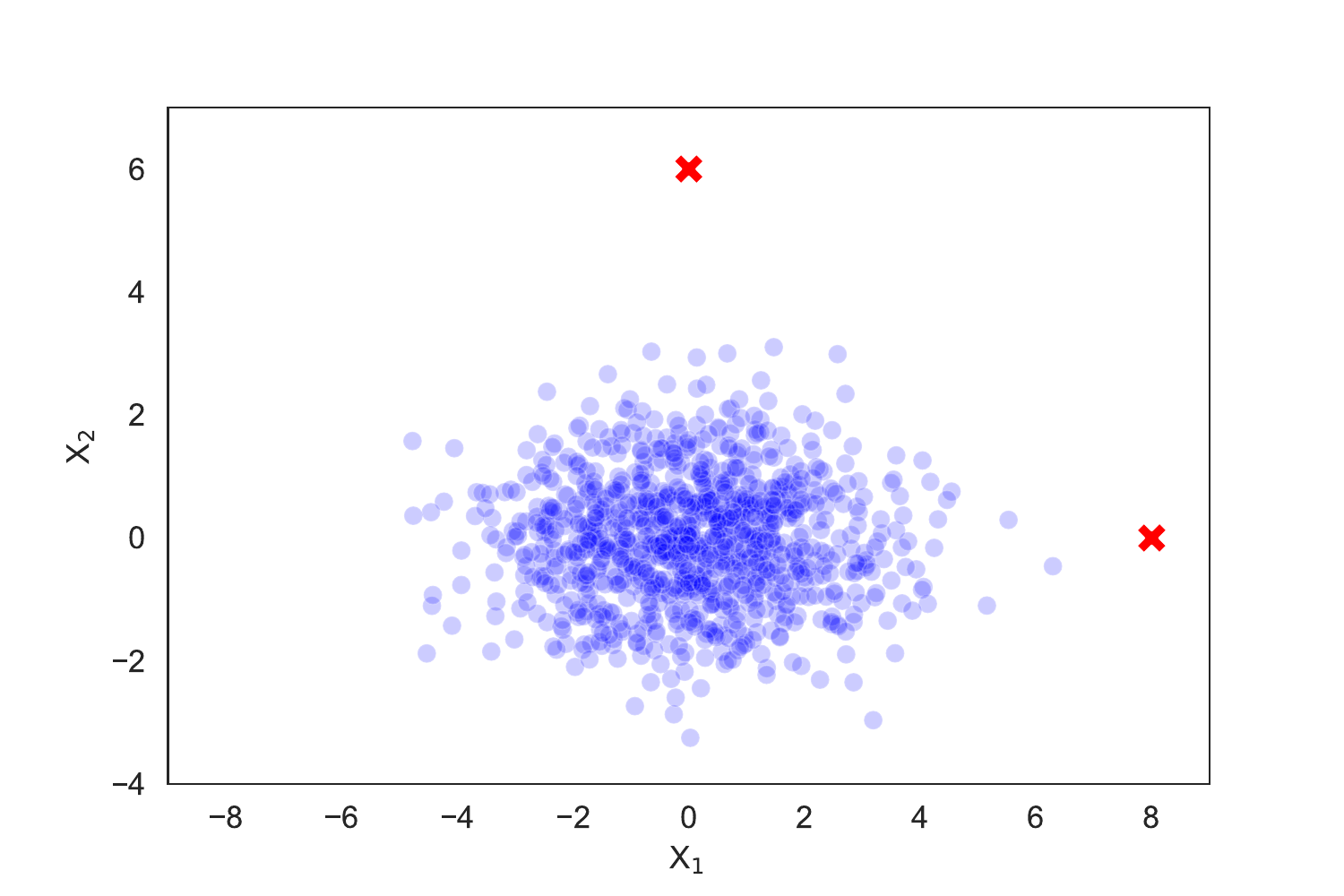}
         \caption{Univariate anomaly example}
         \label{fig:univariate_anomalies}
     \end{subfigure}
     \hfill
     \begin{subfigure}[b]{0.45\textwidth}
         \centering
         \includegraphics[width=\textwidth]{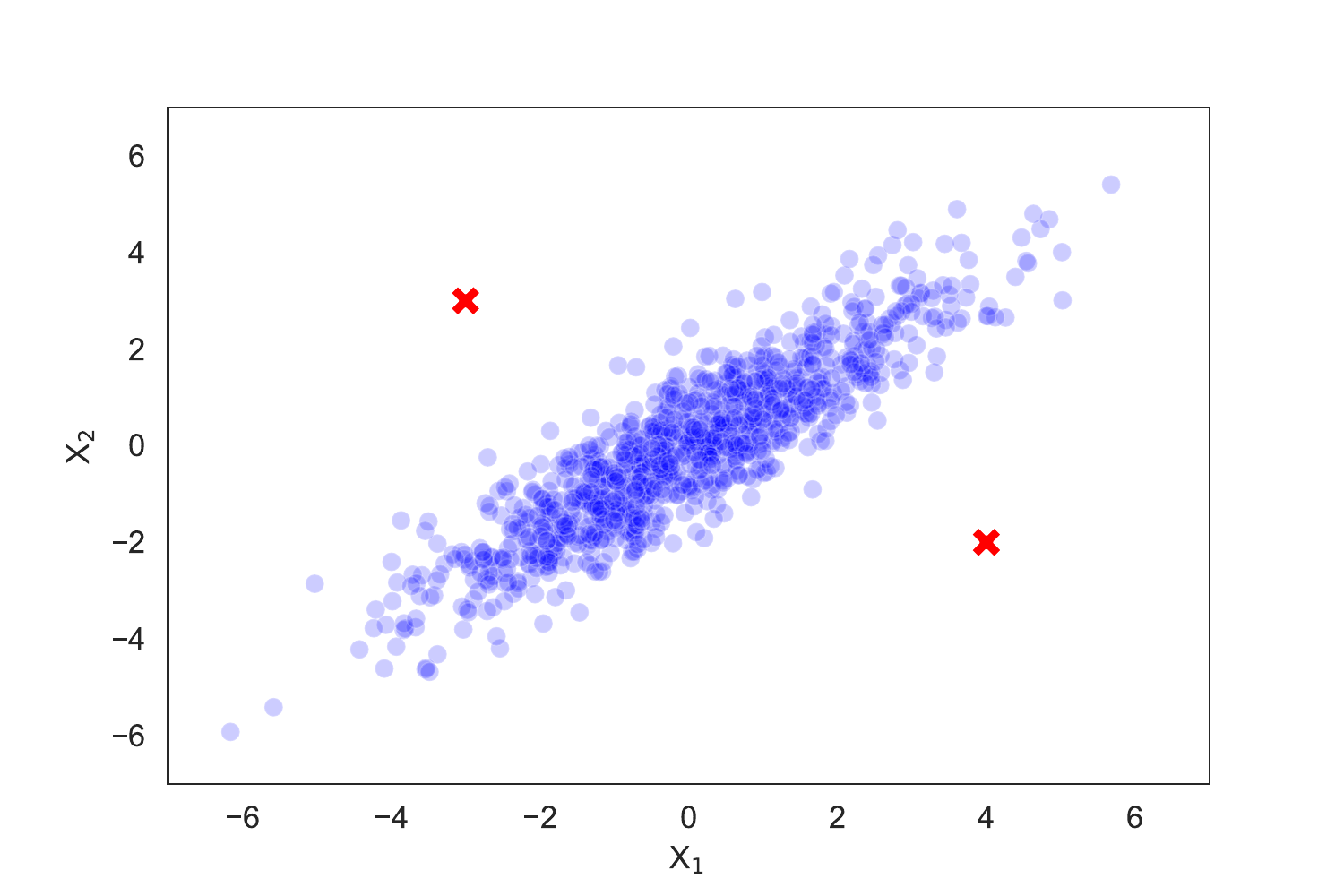}
         \caption{Multivariate anomaly example}
         \label{fig:multivariate_anomalies}
     \end{subfigure}
        \caption{8 examples of different types of anomalies along the 4 defined property axes. Normal data are visualized as blue points, while anomalies are visualized as red crosses.}
        \label{fig:anomaly examples}
\end{figure}

\subsubsection{Enclosed and peripheral anomalies}
Anomalies can be surrounded in the feature space by normal data. When this occurs, we define them as enclosed anomalies. On the other end of this axis, peripheral anomalies occupy the edges of the feature space, and have one or more attribute scores either below the minimum or above the maximum of the scores of the normal data region. Examples of both enclosed and peripheral anomalies can be found in Figures \ref{fig:enclosed_anomalies} and \ref{fig:peripheral_anomalies} respectively.

\subsubsection{Global and local density anomalies}
The most commonly discussed types of anomalies are the global and local anomalies. These definitions follow from the work of \citet{breunig2000lof}. Global anomalies are points which can be isolated from normal data because they occupy a lower density region of the feature space. Local anomalies however, cannot be separated using just these criteria. Local anomalies rather are located in regions with a density which is low compared to nearby normal regions. This allows for well-defined anomalies even when multiple clusters with differing density functions exist in the feature space. Examples of both global and local density anomalies can be found in Figures \ref{fig:global_anomalies} and \ref{fig:local_anomalies} respectively.

\subsubsection{Isolated and clustered anomalies}
Most often, anomalies are isolated, and consist of single datapoints without any additional, normal or anomaly, datapoints nearby. In many practical cases, anomalies are not that singular, and small groups of anomalies form clusters, leading to clustered anomalies. Clustered anomalies are closely related to the phenomenon known as ``masking", where similar anomalies mask each other's presence by forming a cluster~\citep{liu2008isolation}. Examples of both isolated and clustered anomalies can be found in Figures \ref{fig:isolated_anomalies} and \ref{fig:clustered_anomalies} respectively.

\subsubsection{Univariate and multivariate anomalies}
Some anomalies are clearly univariate in nature. That is, they can be identified by just a single feature score in an anomalous range. Other anomalies are multivariate in nature, requiring a specific combination of feature scores to be identified as anomalies. These multivariate anomalies are also often called dependency anomalies, as they differ from the normal dependency, or causal, structure of the data. Examples of both isolated and clustered anomalies can be found in Figures \ref{fig:univariate_anomalies} and \ref{fig:multivariate_anomalies} respectively.

\section{Materials and Methods}
We evaluate the effectiveness of \nalgorithms different algorithms, listed in Table \ref{table:algorithms}.
We evaluate each algorithm multiple times for each dataset, each with a different set of sensible hyperparameters. The results are then averaged across hyperparameter settings, leading to a single average ROC-AUC score for each method-dataset combination. We refrain from optimizing hyperparameters, e.g., using cross-validation, to reflect the real-world situation in which no labels are available for training the models. While unsupervised optimization of hyperparameters has been studied \citet{thomas2016learning} it has not been applied to most algorithms considered in this study.

\subsection{Algorithms}

Of the \nalgorithms methods, \npyodalgorithms were used as implemented in the popular Python library for anomaly detection, PyOD~\citep{zhao2019pyod}. As part of this research, we made several contributions to this open source library, such as a memory-efficient implementation of the COF (Connectivity-based Outlier Factor) method, as well as an implementation of the \citet{birge2006many} method for histogram size selection, which is used in HBOS and LODA (lightweight on-line detector of anomalies). For EIF (extended isolation forest) we used the implementation provided by the authors in the Python package ``eif" by \citet{hariri2019extended}. We implemented the ODIN (Outlier Detection using Indegree Number) method in Python, and it is being prepared as a submission to the PyOD package. The ensemble-LOF method was implemented using the base LOF algorithm from PyOD. DeepSVDD is applied based on the publicly available code by its author Lukas Ruff\footnote{\href{https://github.com/lukasruff/Deep-SVDD-PyTorch}{https://github.com/lukasruff/Deep-SVDD-PyTorch}}, and we modified it to work on general tabular datasets.

We left out several of the implemented methods in the PyOD package, such as LOCI and ROD, because they have a time or memory complexity of $\mathcal{O}(n^3)$, with $n$ the number of data points. The PyOD SOS method was also ignored, due to its $\mathcal{O}(n^2)$ memory requirement. None of these methods performed notably well compared to the other algorithms that we included on the smaller datasets where evaluation was feasible.

\newcommand{\boven}[2]{#1_{\textrm{\tiny #2}}}

\begin{table*}

\begin{adjustwidth}{-2.5 cm}{-2.5 cm}\centering\begin{threeparttable}[!htb]
\caption{Overview of the algorithms, the setting of the hyperparameters, the year of original publication, and the author(s). For the neural networks, the ``shrinkage factor" hyperparameter indicates that any subsequent layer in the encoder is defined by: $\textrm{layer size}_{n+1} = \textrm{layer size}_{n} \times \textrm{shrinkage factor}$.}
\scriptsize
\label{table:algorithms}
\begin{tabular}{lrr}\toprule
\textbf{Name} &\textbf{Hyperparameters} &\textbf{Publication}\\\midrule
\textbf{ABOD}& FastABOD, $k=60$ &\cite{kriegel2008angle} \\
\textbf{AE}& $\boven{n}{layers}={1,2,3}$, shrinkage factor$={0.2,0.3,0.5}$ &\cite{japkowicz1995novelty} \\
\textbf{ALAD}& $\boven{n}{layers}=3$, shrinkage factor$={0.2,0.3,0.5}$ &\cite{zenati2018adversarially} \\
\textbf{CBLOF}&$k={2,..,14}$, $\alpha={0.7,0.8,0.9}$, $\beta={3,5,7}$ &\cite{he2003discovering} \\
\textbf{COF} &$k={5,10,15,20,25,30}$ &\cite{tang2002enhancing} \\
\textbf{COPOD}& &\cite{li2020copod}\\
\textbf{DeepSVDD}& $\boven{n}{layers}={1,2,3}$, shrinkage factor$={0.2,0.3,0.5}$ &\cite{ruff2019deep} \\
\textbf{ECOD} & &\cite{li2022ecod}\\
\textbf{EIF} &$\boven{n}{trees}=1000$, $\boven{n}{samples}={128,256,512,1024}$, no replacement, extension levels: ${1,2,3}$&\cite{hariri2019extended} \\
\textbf{ensemble-LOF}&maximum LOF score over $k={5,...,30}$ &\cite{breunig2000lof} \\
\textbf{gen2out}&  &\cite{lee2021gen}\\
\textbf{GMM}& $\boven{n}{gaussians}={1,...,15}$ &\cite{agarwal2007detecting}\\
\textbf{HBOS}& $\boven{n}{bins}$ based on Birgé-Rozenholc criterium &\cite{goldstein2012histogram} \\
\textbf{IF} & $\boven{n}{trees}=1000$, $\boven{n}{samples}={128,256,512,1024}$, no replacement &\cite{liu2008isolation} \\
\textbf{INNE} & 200 estimators &\cite{bandaragoda2018isolation} \\
\textbf{KDE}& Gaussian kernel &\cite{latecki2007outlier} \\
\textbf{kNN}&$k={5,...30}$, mean distance &\cite{ramaswamy2000efficient} \\
\textbf{kth-NN}&$k={5,...30}$, largest distance &\cite{ramaswamy2000efficient} \\
\textbf{LMDD}&$\boven{n}{shuffles}=100$, MAD dissimilarity function &\cite{arning1996linear} \\
\textbf{LODA}&$\boven{n}{bins}$ based on Birgé-Rozenblac criterium, 100 random projections &\cite{pevny2016loda} \\
\textbf{LOF}&$k={5,...,30}$ &\cite{breunig2000lof} \\
\textbf{LUNAR}&$k={5,10,15,20,25,30}$ &\cite{goodge2022lunar} \\
\textbf{MCD}&subset fraction$={0.6,0.7,0.8,0.9}$ &\cite{rousseeuw1999fast} \\
\textbf{OCSVM}&RBF kernel, $\nu={0.5,0.6,0.7,0.8,0.9}$, $\gamma=1/d$ &\cite{scholkopf1999support} \\
\textbf{ODIN}&$k={5,...30}$ &\cite{hautamaki2004outlier} \\
\textbf{PCA}&selected PCs explain $>{30,50,70,90}$\% of variance &\cite{shyu2003novel} \\
\textbf{sb-DeepSVDD}& $\boven{n}{layers}={1,2,3}$, shrinkage factor$={0.2,0.3,0.5}$ &\cite{ruff2019deep} \\
\textbf{SOD}&$k={20,25,30}$, $l={10,14,18}$, $\alpha={0.7,0.8,0.9}$ &\cite{kriegel2009outlier} \\
\textbf{SO-GAAL}& stop epochs$=50$&\cite{liu2019generative} \\

\textbf{u-CBLOF}&$k={2,..,14}$, $\alpha={0.7,0.8,0.9}$, $\beta={3,5,7}$&\cite{amer2012nearest} \\

\textbf{VAE}& $\boven{n}{layers}={1,2,3}$, shrinkage factor$={0.2,0.3,0.5}$ &\cite{an2015variational} \\
\textbf{$\beta$-VAE}& $\boven{n}{layers}={1,2,3}$, shrinkage factor$={0.2,0.3,0.5}$, $\gamma={10,20,50}$ &\cite{zhou2020unsupervised} \\

\bottomrule
\end{tabular}
\end{threeparttable}\end{adjustwidth}
\end{table*}

\subsection{Data}
\subsubsection{Datasets}
In this study we consider a large collection of datasets from a variety of sources. We focus on real-valued, multivariate, tabular data, comparable to the datasets used by \citet{fernandez2014we, campos2016evaluation, goldstein2016comparative, soenen2021effect, domingues2018comparative}. Table \ref{table:datasets} contains a summary of the datasets, listing each dataset's origin, number of samples, number of variables, number and percentage of anomalies, and how many duplicates were removed in the processing of the data. 

Our collection consists for the most part of datasets from the ODDS data collection~\citep{rayana2016odds}, specifically the multi-dimensional point datasets. It is a collection of various datasets, mostly adapted from the UCI machine learning repository~\citep{Dua2019}. All datasets are real-valued, without any categorical data. Curation of this collection is sadly not fully up-to-date, causing some of the listed datasets to be unavailable. The unavailable datasets were omitted from this comparison.

In addition to the ODDS dataset, we also incorporate publicly available datasets used in earlier anomaly detection research: several datasets from the comparison by \citet{goldstein2016comparative}, from the comparison of \citet{campos2016evaluation} using ELKI~\citep{ELKI}, from a study on Generative Adversial Active Learning, or GAAL~\citep{liu2019generative}, from a study on extended Autoencoders~\citep{shin2020extended}, from the ADBench comparison~\citep{han2022adbench}, and from a study on Efficient Online Anomaly Detection (EOAD)~\citep{brandsaeter2019efficient}. Datasets from these latter sources that are (near-)duplicates of datasets present in the ODDS collection are left out. In Table \ref{table:datasets} we specify exactly where each dataset was downloaded or reconstructed from.

\citet{emmott2013systematic, emmott2015meta} present a systematic methodology to construct anomaly detection benchmarks, which is then also extensively applied by \citet{pevny2016loda}. In this paper, we chose not to construct our own benchmark datasets, which inevitably leads to some arbitrariness and possibly bias, but instead we rely on a large collection of different datasets used in earlier comparison studies. Synthetic datasets are not included in this study, as real-world datasets are generally considered the best available tool for benchmarking algorithms~\citep{emmott2015meta, domingues2018comparative, ruff2021unifying}. While real-world datasets are preferred for benchmarking, we note the usefulness of synthetic data in when studying specific properties of anomaly detection algorithms. 

\begin{table*}
\caption{Summary of the \ndatasets multivariate datasets used in our anomaly detection algorithm comparison: the colloquial name of the dataset, origin of the dataset, the number of samples after removal of duplicates, variables, anomalies and removed duplicates, as well as the percentage of anomalies.}

\label{table:datasets}
\centering

\resizebox{\columnwidth}{!}{%
\input{Tables/datasets_table.tex}
}
\end{table*}

\subsubsection{Preprocessing}

Several steps have been undertaken to be able to compare the performance of the various algorithms on the different datasets. Firstly, all duplicate samples have been removed from each dataset, as many anomaly detection methods cannot handle data with duplicates~\citep{campos2016evaluation}.
Furthermore, all variables in all datasets have been scaled and centered. Although some algorithms, such as Isolation Forest, can implicitly handle variables with different scales, methods that involve, for example, distance or cross-product calculations are strongly affected by the scale of the variables. Centering is done for each variable in a dataset by subtracting the median. Scaling is performed by dividing each variable by its interquartile range. Our choice of centering and scaling procedure is deliberate, as both the median and interquartile range are influenced less by the presence of anomalies than the mean and standard deviation. This procedure is generally considered to be more stable than standardization when anomalies are known to be present~\citep{rousseeuw1993alternatives}.

\subsection{Evaluation Procedure}

Due to the unsupervised nature of anomaly detection, it is generally more useful to evaluate anomaly scores, rather than binary labels as also produced by some algorithms. Using scores, samples can be ranked, providing insights into the underlying nature of anomalies. 
For each dataset, we calculate anomaly scores on all available data at once, without using any cross-validation or train-test splits. The scores from this unsupervised analysis are then used in conjunction with the ground truth labels to evaluate the performance of the algorithm.
In order to compare the different algorithms we calculate the performance for each algorithm-dataset combination in terms of the AUC (area under the curve) value resulting from the ROC (receiver operating characteristic) curve. This is the most commonly used metric in anomaly detection evaluations~\citep{goldstein2016comparative, campos2016evaluation, xu2018comparison}, which can be readily interpreted from a probabilistic view. We considered using other metrics, such as the R-precision or average precision and their chance-adjusted variants introduced by \citet{campos2016evaluation}, but found these to be less stable, and harder to interpret.

For each dataset we rank the AUC scores calculated from the scores produced by each algorithm. Following the recommendations for the comparison of classifiers by \citet{demvsar2006statistical}, we use the Iman-Davenport statistic~\citep{iman1980approximations}  in order to determine whether there is any significant difference between the algorithms. If this statistic falls below the desired critical value corresponding to a $p$-value of 0.05, we apply the Nemenyi post-hoc test~\citep{nemenyi1963distribution} to then assess which algorithms differ significantly from each other. 

In some of the visualizations in this paper we plot the percentage of maximum AUC, defined as
\[
    \widetilde{\textrm{AUC}}(a,d) = \frac{\textrm{AUC}(a,d)}{\max_{a' \in A} \textrm{AUC}(a',d)} \times 100 \: ,
\]
with $a$ one of the $A$ algorithms and $d$ one of the $D$ datasets.

\subsection{Reproducibility}
In order to reproduce all our experiments, we have provided access to a public GitHub repository\footnote{\href{https://github.com/RoelBouman/outlierdetection}{https://github.com/RoelBouman/outlierdetection}} containing the code and datasets used for all experiments as well as for the production of all figures and tables presented in this paper. 

\section{Results}

\subsection{Overall algorithm performance}
\label{subsection:overall_performance}
In order to gauge the performance, we evaluated each algorithm on each dataset using the AUC measure corresponding to a ROC curve. To evaluate the performance across multiple sensible hyperparameters the AUC value for a given method is the average of the AUC of each hyperparameter setting evaluated.
In our analysis, we found three datasets on which nearly every evaluated algorithm performed subpar, i.e. with all AUC values between 0.4 and 0.6. These datasets, the `hrss\_anomalous\_standard', `wpbc' and `yeast' dataset, were therefore excluded from further analysis. Some datasets showed no AUC values above 0.6, but did show AUC values below 0.4. In these cases, the detector performs better when the labels are inverted. This behaviour was observed in the `skin' and `vertebral' datasets. The construction of these datasets was done based on treating the largest group of samples as the normal (0) class, and the smaller group as the anomaly (1) class. Yet, for both these sets, the more heterogeneous group is chosen as the normal class, in constrast to normal anomaly definitions. For these datasets, we inverted the labelling and recalculated the AUC values.

\begin{figure}
\includegraphics[width=\textwidth]{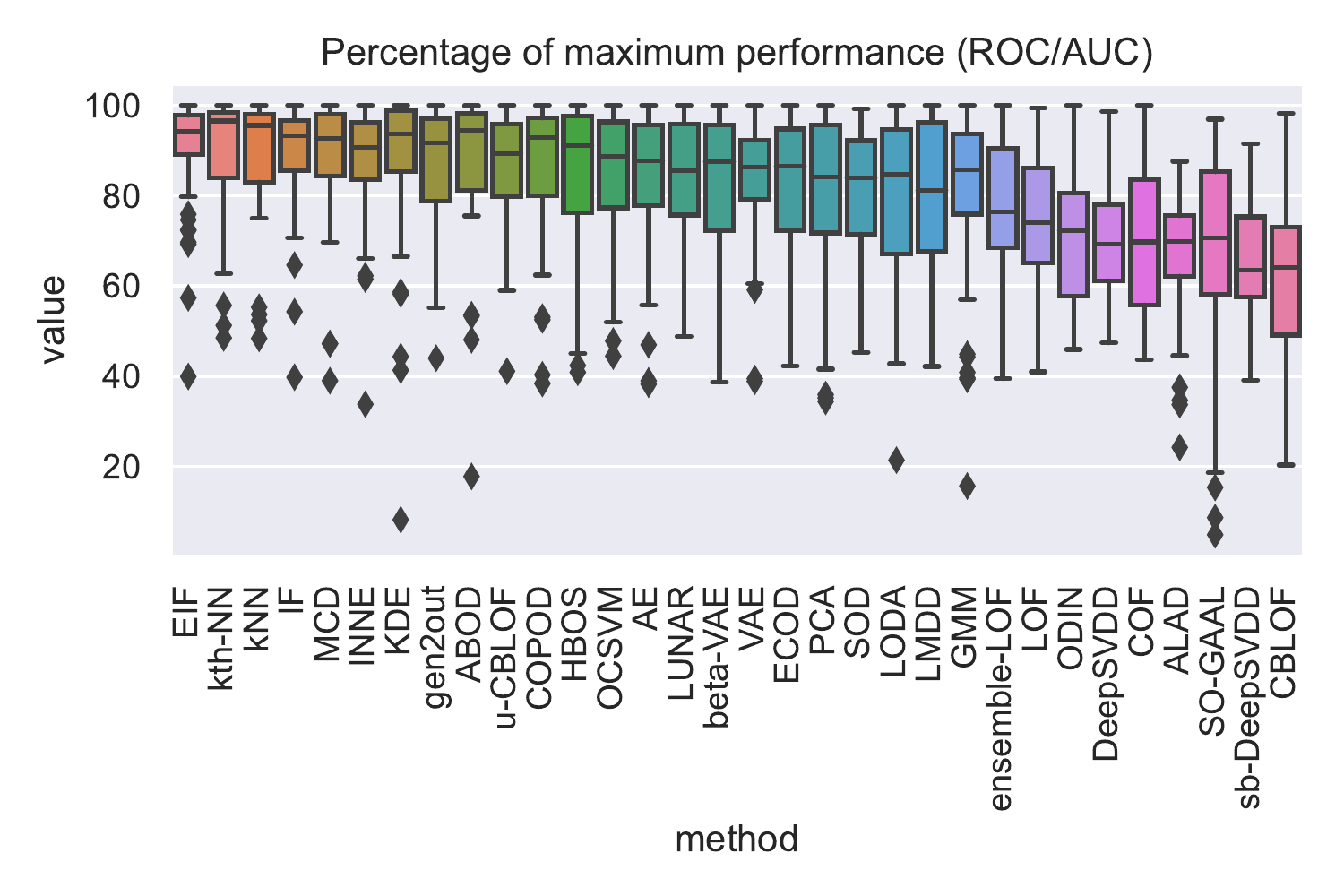}
\caption{Boxplots of the performance of each algorithm on each dataset in terms of percentage of maximum AUC. The maximum AUC is the highest AUC value obtained by the best performing algorithm on that particular dataset. The whiskers in the boxplots extend 1.5 times the interquartile range past the low and high quartiles. Dataset-algorithm combinations outside of the whiskers are marked as diamonds.} 
\label{fig:ROCAUC_boxplot_all_datasets}
\end{figure}

Figure \ref{fig:ROCAUC_boxplot_all_datasets} shows the distribution of the performance for each method. In order to compare the AUC across different datasets, which might have different baseline performances, in a boxplot we express the AUC in terms of its percentage of the maximum AUC value obtained by the best performing algorithm on that particular dataset. 

It can be seen that many of the algorithms perform comparably, with a median percentage of maximum AUC around 90\%. Several lower medians, as well as wider quartiles, can be observed. 

To determine whether any of the observed differences in performance from Figure~\ref{fig:ROCAUC_boxplot_all_datasets} are significant, we apply the Iman-Davenport test. This yielded a test-statistic of \imamdavenportoverall, far above the critical value of \imamdavenportcriticaloverall, thus refuting the null hypothesis. We then applied the Nemenyi post-hoc test to establish which algorithms significantly outperform which other algorithms. The results of this Nemenyi post-hoc test are summarized in Table~\ref{table:nemenyi_summary_all_datasets}.

\begin{table*}
\caption{Significant differences between algorithms based on Nemenyi post-hoc analysis. ++/+ denotes that the row algorithm outperforms the column algorithm at $p=0.05$/$p=0.10$ respectively, while -{}- denotes that the row algorithm is outperformed by the column algorithm at $p=0.05$. Rows and columns are sorted by descending and ascending mean performance, respectively. Columns are not shown when the column algorithm is not outperformed by any other algorithms at $p=0.05$ or $p=0.10$. The last column shows the mean AUC.}

\label{table:nemenyi_summary_all_datasets}
\centering
\resizebox{\columnwidth}{!}{%
\input{Tables/nemenyi_summary_truncated}
}
\end{table*}

Table~\ref{table:nemenyi_summary_all_datasets} reveals that there are indeed several algorithms significantly outperforming many other algorithms. Most notable here is $k$th-NN, which significantly outperforms 14/15 algorithms evaluated in this study at the $p=0.05/p=0.10$ significance level respectively. 
The popular Isolation Forest method and its extended version show similar consistent performance. Since the computational complexity of Isolation Forest and variants thereof scales linearly with the number of samples $n$, this may give them a clear edge over methods such as $k$NN and derivatives for large datasets, with a computational complexity that scales quadratically or at best with $\mathcal{O}(n\log{n})$ when optimized. 

From Figure \ref{fig:ROCAUC_boxplot_all_datasets} and Table \ref{table:nemenyi_summary_all_datasets} we can also observe that the original CBLOF method is by far the worst performing method, being significantly outperformed by 21/22 algorithms at at the $p=0.05/p=0.10$ significance level respectively.. This corroborates the results of \citet{goldstein2016comparative}, who also found CBLOF to consistently underperform, while its unweighted variant, u-CBLOF, performs comparably to other algorithms. 

From these overall results it is clear that many of the neural networks do not perform well. (Soft-boundary) DeepSVDD, ALAD, and SO-GAAL all occupy the lower segment of overall method performance. We surmise that there are two likely reasons for this phenomenon. Firstly, these methods were not designed with tabular data in mind, and they can't leverage the same feature extraction capabilities that give them an edge on their typical computer vision tasks. Secondly, these methods are relatively complex, making it exceedingly hard to specify general hyperparameter settings and architectures which work on a large variety of datasets.
Not all neural networks suffer from this problem, as the auto-encoder and variants, as well as LUNAR, perform about average. This is likely caused by more straightforward optimisation criteria.

In addition to the neural networks, the local methods, such as LOF, ODIN, COF, and CBLOF, are some of the most underperforming methods. This result for LOF stands in stark contrast to the results of \citet{campos2016evaluation}, who found LOF to be among the best performing methods. This is most likely caused by their evaluation on a small number of datasets with a low percentage of anomalies, which causes LOF to suffer less from swamping or masking~\citep{liu2008isolation}. We further study this finding in Section \ref{subsection:local_global}.

\subsection{Clustering algorithms and datasets}
\label{subsection:similarities}

\begin{figure*}
\includegraphics[width=\textwidth]{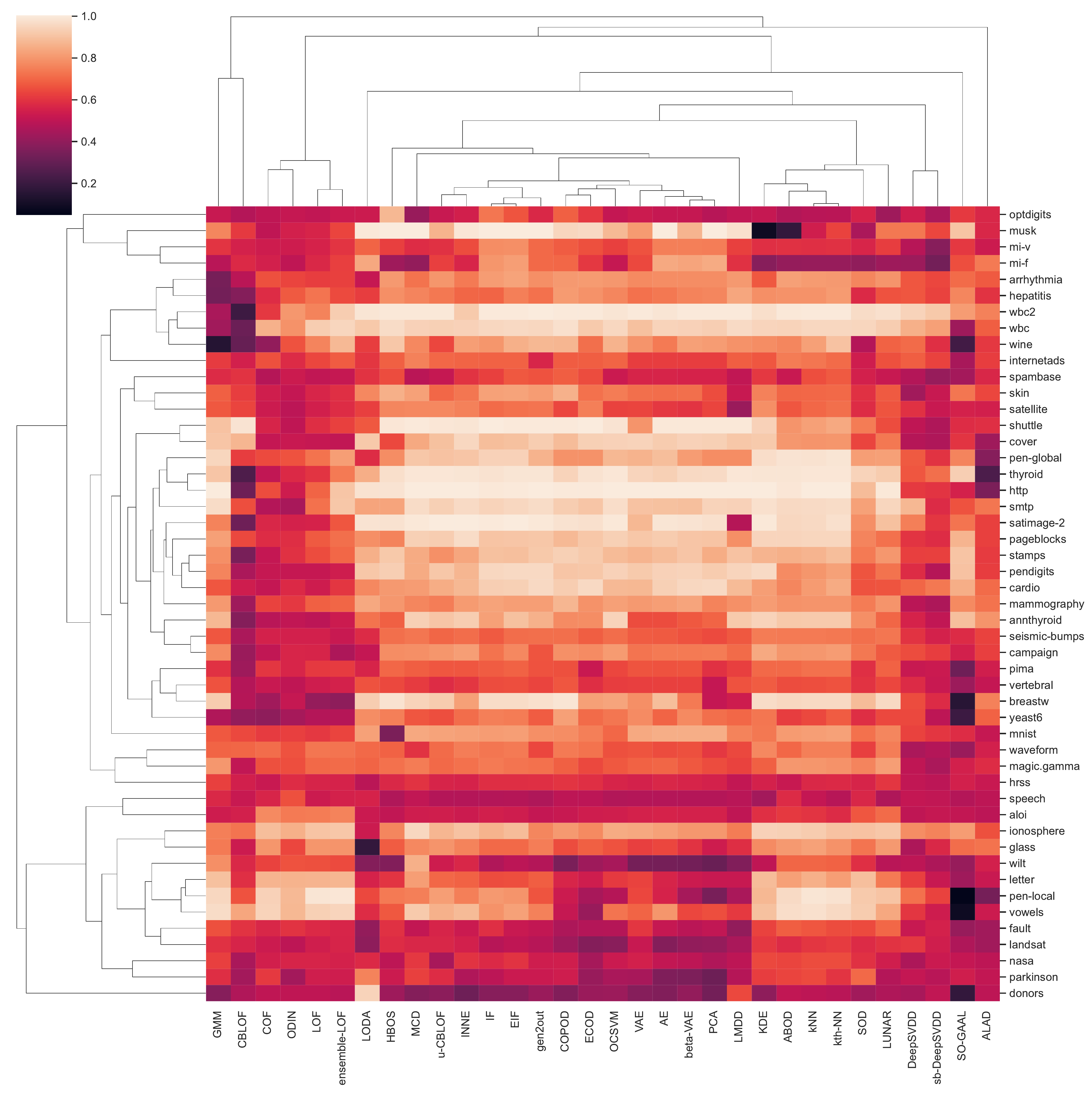}
\caption{Clustered heatmap of the ROC/AUC performance of each algorithm. The algorithms and datasets are each clustered using hierarchical clustering with average linkage and the Pearson correlation as metric.}
\label{figure:clustermap}
\end{figure*}

To visualize the similarities between algorithms on one hand, and the datasets on the other, Figure~\ref{figure:clustermap} shows a heatmap of the performance of each dataset/algorithm combination and dendrograms of two hierarchical clusterings, one on the datasets, and one on the algorithms. We applied average linkage cluster analysis with Pearson correlation as distance measure and optimized the leaf orderings for visualization with the method of \citet{bar2001fast}.

Figure~\ref{figure:clustermap} shows that many similar algorithms cluster together in an expected way, with families of algorithms forming their own clusters. Some interesting patterns can be observed at a larger level. For the algorithms, we obtain several fairly distinct clusters. Firstly, CBLOF is distinct, as it underperforms on nearly every dataset. Similarly, underperforming methods such as SO-GAAL, GMM, and ALAD do not clearly belong to other emerging clusters.
The local methods, COF, ensemble-LOF, LOF and ODIN, form a separate cluster. These algorithms, which are specifically designed to detect local anomalies, work well on a few (approximately a quarter) of the datasets, but do not perform well for most other datasets. We have a small cluster of $k$NN and related methods such as ABOD and SOD, that performs decently for all datasets. Lastly, a large cluster of methods seems to negatively correlate with the local methods, performing well for most (approximately three-quarters) of the datasets, but less so for the remainder.

The datasets split into two clearly distinct clusters: one cluster of datasets on which the local algorithms perform well, and another cluster of datasets on which the large cluster of algorithms performs well. Combining the two-way clustering with knowledge of the algorithms suggests that approximately one quarter of the datasets comprises so-called local problems, while the other three-quarters comprises global problems. This is corroborated by specifically constructed local and global sets `pen-local' and `pen-global', that clearly belong to their expected clusters. This observation is corroborated by research by \citet{steinbuss2021benchmarking} and \citet{emmott2015meta}, who similarly find differences between what they categorize as local/dependency and multi-cluster anomalies respectively, and global anomalies. We can not clearly observe any other clear patterns of different anomaly properties arising from our analysis.

To the best of our knowledge, no previous study on naturally occurring anomalies in real-world data has looked, in detail, into the difference between the performance of algorithms when specifically being applied to either global or local anomaly detection problems.

\subsection{Performance on global and local problems}
\label{subsection:local_global}
In the previous section, we discovered a clear distinction between two clusters of datasets: one with the ``local'' datasets `aloi', `donors', `fault', `glass', `ionosphere', `landsat', `letter', `nasa', `parkinson', `pen-local', `vowels', and `wilt', and another with the remaining ``global'' datasets. Suspecting that different methods may do well on different types of datasets, we repeated the significance testing procedure from Section \ref{subsection:overall_performance} for both clusters separately. 

\begin{figure}
\includegraphics[width=\textwidth]{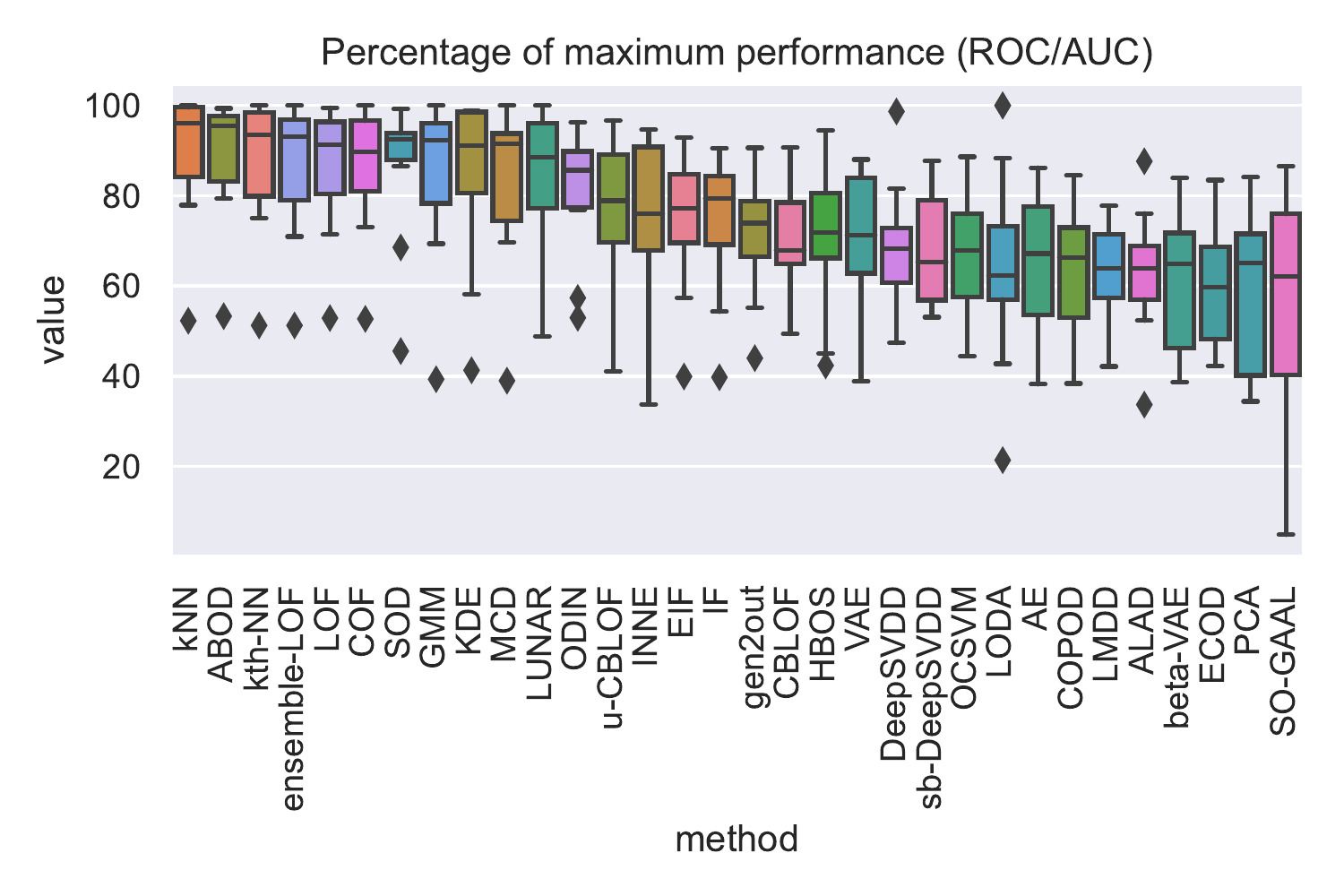}
\caption{Boxplots of the performance of each algorithm on the ``local" datasets in terms of percentage of maximum AUC. The maximum AUC is the highest AUC value obtained by the best performing algorithm on that particular dataset. The whiskers in the boxplots extend 1.5 times the interquartile range past the low and high quartiles. Dataset-algorithm combinations outside of the whiskers are marked as diamonds.} 
\label{fig:ROCAUC_boxplot_local_datasets}
\end{figure}

Performance boxplots for all algorithms applied on the collection of local datasets can be found in Figure \ref{fig:ROCAUC_boxplot_local_datasets}.
Figure~\ref{fig:ROCAUC_boxplot_local_datasets} clearly shows the reversed performance of some of the local methods for anomaly detection. Where COF, ensemble-LOF, and LOF were among the worst performers over the entire collection, they are among the best performers when applied to the problems for which they were specifically developed. This phenomenon is a fine example of Simpson's paradox~\citep{simpson1951interpretation}. This also partially explains the difference in findings of our overall comparison and the comparison of \citet{campos2016evaluation}.

We then repeated the Nemeny-Friedman post hoc test on just the local datasets. The results for this analysis are summarized in Table~\ref{table:nemenyi_summary_local_datasets}.
~$k$NN, and ABOD are the top performers, and significantly outperform 16 other methods at $p=0.05$.

\begin{table*}
\caption{Significant differences between algorithms on the collection of local problems based on Nemenyi post-hoc analysis. ++/+ denotes that the row algorithm outperforms the column algorithm at $p=0.05$/$p=0.10$. Rows and columns are sorted by descending and ascending mean performance, respectively. Columns are not shown when the column algorithm is not outperformed by any other algorithms at $p=0.05$ or $p=0.10$. The last column shows the mean AUC.}
\centering
\resizebox{\columnwidth}{!}{%
\input{Tables/nemenyi_summary_local_truncated}
}
\label{table:nemenyi_summary_local_datasets}
\end{table*}

\begin{figure}
\includegraphics[width=\textwidth]{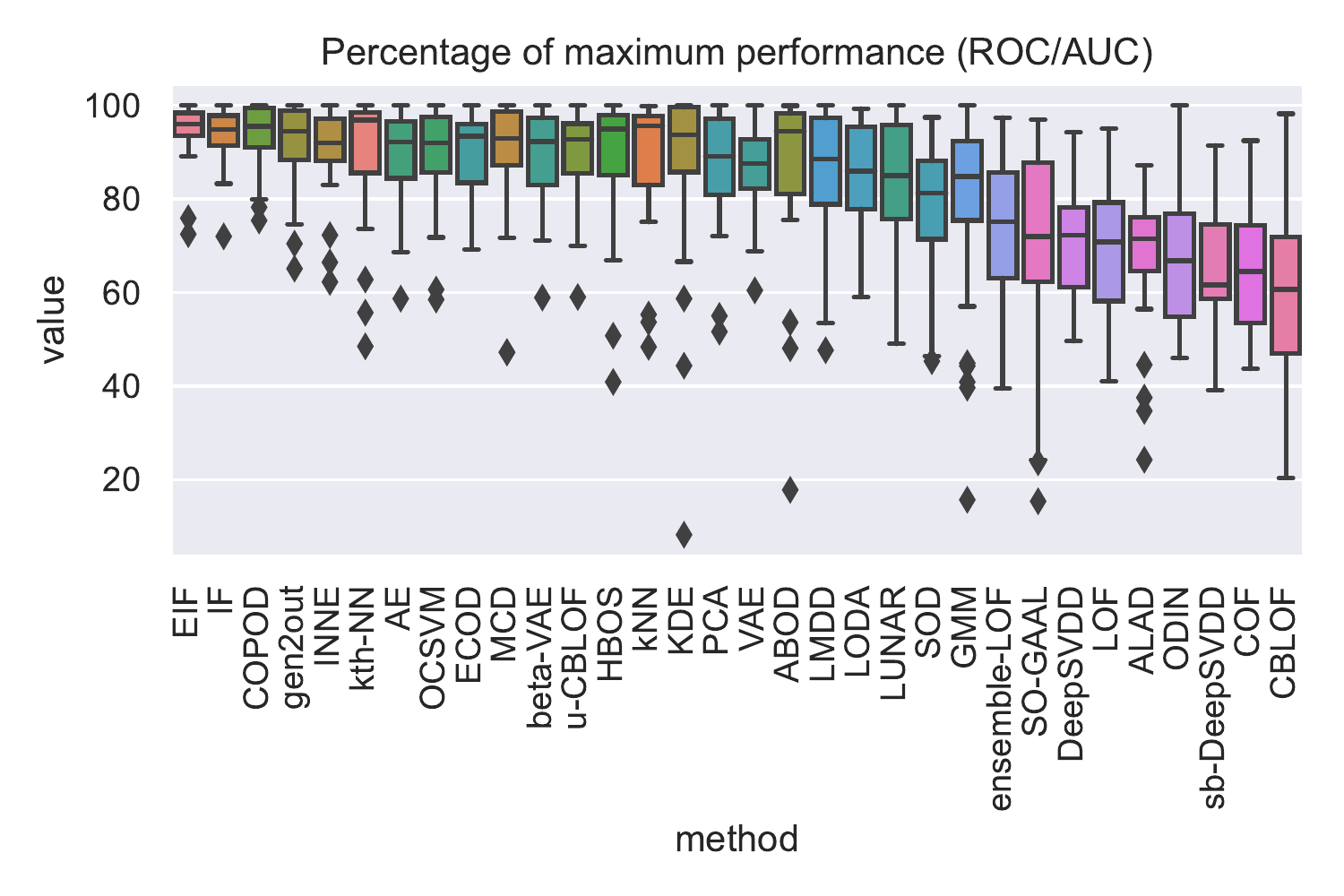}
\caption{Boxplots of the performance of each algorithm on the global datasets in terms of percentage of maximum AUC. The maximum AUC is the highest AUC value obtained by the best performing algorithm on that particular dataset. The whiskers in the boxplots extend 1.5 times the interquartile range past the low and high quartiles. Dataset-algorithm combinations outside of the whiskers are marked as diamonds.} 
\label{fig:ROCAUC_boxplot_global_datasets}
\end{figure}

We then repeated the analysis for the global datasets, leading to the performance boxplots in Figure \ref{fig:ROCAUC_boxplot_global_datasets} and the significance results in Table \ref{table:nemenyi_summary_global_datasets}. 

\begin{table*}
\caption{Significant differences between algorithms on the collection of global problems based on Nemenyi post-hoc analysis. ++/+ denotes that the row algorithm outperforms the column algorithm at $p=0.05$/$p=0.10$. Rows and columns are sorted by descending and ascending mean performance, respectively. Columns are not shown when the column algorithm is not outperformed by any other algorithms at $p=0.05$ or $p=0.10$. The last column shows the mean AUC.}
\centering
\resizebox{\columnwidth}{!}{%
\input{Tables/nemenyi_summary_global_truncated}
}
\label{table:nemenyi_summary_global_datasets}
\end{table*}

From Figure \ref{fig:ROCAUC_boxplot_global_datasets} and Table \ref{table:nemenyi_summary_global_datasets} we can see that the Extended Isolation Forest has the highest mean performance, closely followed by the regular Isolation Forest. The Extended Isolation Forest outperforms 13/15 methods at $p=0.05/p=0.10$ respectively. Coincidentally, these methods also have the lowest computational and memory requirement, leaving them as the most likely choices for global anomalies.

\section{Discussion}
In our study we compared the performance of anomaly detection algorithms on \ndatasets semantically meaningful real-world tabular datasets, more than any other recent comparison studies~\citep{campos2016evaluation, goldstein2016comparative, xu2018comparison, soenen2021effect, steinbuss2021benchmarking, domingues2018comparative, han2022adbench}. A somewhat comparable study by \citet{fernandez2014we} on classification algorithms easily considered 121 datasets. The main reason for this discrepancy is that datasets for comparing anomaly detection algorithms rarely include categorical variables, which are not an issue for comparing classification algorithms. It is certainly possible to further extend the collection of dataset, e.g., through dataset modifications. \citet{campos2016evaluation}, \citet{emmott2013systematic, emmott2015meta}, and \citet{steinbuss2021benchmarking} modified datasets in different ways to create similar datasets with differing characteristics from a single base dataset. While such modifications can be useful for targeted studies, near-duplicate datasets are far from independent and then seem detrimental to a proper statistical comparison of anomaly detection algorithms, such as can be observed in \citet{emmott2015meta}.

In this study we compared \nalgorithms of the most commonly used algorithms for anomaly detection. This collection is certainly not exhaustive: many more methods exist~\citep{ELKI, goldstein2016comparative, emmott2015meta, ruff2021unifying, domingues2018comparative}, and likely even more will be invented. Also along this axis, there is a clear discrepancy with the study by \citet{fernandez2014we} on classification algorithms, who incorporated 179 classifiers from 17 different families. Apparently, the number of classification algorithms largely exceeds the number of anomaly detection algorithms. But perhaps more importantly, there are many more solid and easy-to-use implementations of classification algorithms in many different machine learning libraries than there are out-of-the-box implementations of anomaly detection algorithms. Before being able to perform the comparison in this study, we had to spend quite some effort to clean up and sometimes re-implement (parts of) existing code.

In this research we chose not to cover meta-techniques for ensembling. While ensembles are of great interest, a better understanding of the performance of base learners is an essential prerequisite before moving on to a study of ensemble methods. 

While we evaluated neural networks in our comparison, no general guidelines exist on how to construct a well-performing network for any given dataset, which is essential for the unsupervised setup considered in this study. Additionally, the strength of many of these methods comes from high-level feature extraction implicitly performed by the network, which cannot be leveraged on the smaller tabular datasets used in this benchmark. Like \citet{ruff2021unifying}, we recognize that there is a major discrepancy between the availability of classification and anomaly detection benchmark datasets useful for deep learning approaches. More anomaly detection benchmark datasets useful for deep learning based anomaly detection would be a welcome addition to the field.

Cross-comparing the performance of algorithms on datasets, we noticed a clear separation between two clusters of datasets and roughly two clusters of corresponding algorithms. We characterized these clusters as ``local'' and ``global'' datasets and algorithms, in correspondence with common nomenclature in the literature~\citep{breunig2000lof, goldstein2016comparative}. However, we are well aware that this characterization may turn out to be an oversimplification when analyzing more datasets and more algorithms in closer detail. For example, the local and global problems likely have quite some overlap, but need not be fully equivalent with multimodal and unimodal anomaly detection problems, respectively. Overlap between multimodal and local problems occurs when the different modes start having different densities, so that local algorithms that try to estimate these local densities fare better than global algorithms that cannot make this distinction. Further theoretical and empirical studies, e.g., on carefully simulated datasets are needed to shed further light.

\section{Conclusion}
Based on our research we can establish general guidelines on when users should apply which anomaly detection methods for their problem.

In general, when a user has no \textit{a priori} knowledge on whether or not their dataset contains local or global anomalies, $k$-thNN is the best choice. It outperforms 14 out of \nalgorithms other evaluated methods at $p=0.05$, and is one the highest performing method based on its mean AUC score. In case the $\mathcal{O}(n^2)$ (or $\mathcal{O}(n \log{n})$ when optimized) computational complexity of $k$NN is an issue, the slightly worse performing Isolation Forest or its extended variant, are good alternatives.

When a dataset is known to contain local anomalies, the best performing method is $k$NN, which outperforms 16 out of \nalgorithms methods at $p=0.10$. 

Datasets containing just global anomalies are best analyzed using IF (isolation forest) or, preferably, its extended variant EIF, which is the top performing algorithm on the datasets containing global anomalies. COPOD, gen2out, INNE and $k$-thNN all perform comparably, and these methods all outperform at least 10 other methods at $p=0.05$. IF and EIF are the algorithms with the lowest computational complexity, which are usually preferable in practice.

Contemplating the above considerations, we are tempted to answer the question in the title of our paper with ``three'': a toolbox with $k$-thNN, $k$NN, and EIF seems sufficient to perform well on the type of multivariate datasets considered in our study. These three algorithms are due to the scope of this study likely to perform well on unseen real-world multivariate datasets. This conclusion is open for further consideration when other algorithms and/or datasets are added to the bag, which should be relatively easy to check when extending the code and the dataset pre-processing procedures that we open-sourced. Future work following this study may seek to extend our comparative analysis with diverse types of data such as raw images, texts and time-series.

\acks{The research reported in this paper has been partly funded by the NWO grant NWA.1160.18.238 (\href{https://primavera-project.com/}{PrimaVera})}, as well as BMK, BMDW, and the State of Upper
Austria in the frame of the SCCH competence center INTEGRATE [(FFG grant no. 892418)] part of the FFG COMET Competence Centers for Excellent Technologies Programme.

\newpage

\appendix

\section{AUC scores for each algorithm-dataset combination}
See Table \ref{table:AUC_all_datasets}.

\begin{table*}

\caption{The AUC values for each algorithm-dataset combination.}
\label{table:AUC_all_datasets}

\let\center\empty
\let\endcenter\relax
\centering

\begin{adjustbox}{angle=90}

\resizebox{0.75\pdfpageheight}{!}{\input{Tables/AUC_all_datasets}}
\end{adjustbox}
\end{table*}

\section{Nemenyi post-hoc analysis results}
Table \ref{table:nemenyi_table_all} shows the results for all datasets, Table \ref{table:nemenyi_table_local} shows the results for the local datasets, and Table \ref{table:nemenyi_table_global} shows the results for the global datasets.

\begin{table*}
\caption{The p-values from Nemenyi post-hoc analysis on all algorithm pairs based on all \ndatasets datasets. P-values below 0.05 have been printed bold.}
\label{table:nemenyi_table_all}
\let\center\empty
\let\endcenter\relax
\centering
\resizebox{\columnwidth}{!}{\input{Tables/nemenyi_table_all_datasets}}
\end{table*}

\begin{table*}
\caption{The p-values from Nemenyi post-hoc analysis on all algorithm pairs based on the \nlocaldatasets local datasets. P-values below 0.05 have been printed bold.}
\label{table:nemenyi_table_local}
\let\center\empty
\let\endcenter\relax
\centering
\resizebox{\columnwidth}{!}{\input{Tables/nemenyi_table_local}}
\end{table*}

\begin{table*}
\caption{The p-values from Nemenyi post-hoc analysis on all algorithm pairs based on the \nglobaldatasets global datasets. P-values below  0.05 have been printed bold.}
\label{table:nemenyi_table_global}
\let\center\empty
\let\endcenter\relax
\centering
\resizebox{\columnwidth}{!}{\input{Tables/nemenyi_table_global}}
\end{table*}
\bibliography{ref.bib}

\end{document}

%% file: Tables/datasets_table.tex
\begin{tabular}{llrrrlrr}
\toprule
{} &        Origin &  \#samples &  \#variables &  \#outliers & \%outliers &  \#duplicates &  \#removed variables \\
Name                     &               &           &             &            &           &              &                     \\
\midrule
aloi                     &     Goldstein &     49533 &          27 &       1507 &   (3.04\%) &          466 &                   0 \\
annthyroid               &          ODDS &      7062 &           6 &        534 &   (7.56\%) &          138 &                   0 \\
arrhythmia               &          ODDS &       452 &         257 &         66 &   (14.6\%) &            0 &                  17 \\
backdoor                 &       ADBench &     87020 &         193 &       1879 &   (2.16\%) &         8309 &                   3 \\
breastw                  &       ADBench &       449 &           9 &        236 &  (52.56\%) &          234 &                   0 \\
campaign                 &       ADBench &     41176 &          62 &       4639 &  (11.27\%) &           12 &                   0 \\
cardio                   &          ODDS &      1822 &          21 &        175 &    (9.6\%) &            9 &                   0 \\
celeba                   &  ADRepository &    113983 &          39 &       2911 &   (2.55\%) &        88616 &                   0 \\
cover                    &          ODDS &    286048 &          10 &       2747 &   (0.96\%) &            0 &                   0 \\
donors                   &       ADBench &     35421 &          10 &       7667 &  (21.65\%) &       583905 &                   0 \\
fault                    &       ADBench &      1941 &          27 &        673 &  (34.67\%) &            0 &                   0 \\
fraud                    &       ADBench &    275661 &          29 &        473 &   (0.17\%) &         9146 &                   0 \\
glass                    &          ODDS &       213 &           9 &          9 &   (4.23\%) &            1 &                   0 \\
hepatitis                &          ELKI &        80 &          19 &         13 &  (16.25\%) &            0 &                   0 \\
hrss\_anomalous\_optimized &         ex-AE &      5874 &          18 &       1468 &  (24.99\%) &        13760 &                   0 \\
hrss\_anomalous\_standard  &         ex-AE &      7040 &          18 &       1830 &  (25.99\%) &        16605 &                   0 \\
http                     &          ODDS &    221900 &           3 &         72 &   (0.03\%) &       345598 &                   0 \\
internetads              &          ELKI &      1966 &        1555 &        368 &  (18.72\%) &            0 &                   0 \\
ionosphere               &          ODDS &       350 &          33 &        125 &  (35.71\%) &            1 &                   0 \\
landsat                  &       ADBench &      6435 &          36 &       1333 &  (20.71\%) &            0 &                   0 \\
letter                   &          ODDS &      1598 &          32 &        100 &   (6.26\%) &            2 &                   0 \\
magic.gamma              &       ADBench &     18905 &          10 &       6573 &  (34.77\%) &          115 &                   0 \\
mammography              &          ODDS &      7848 &           6 &        253 &   (3.22\%) &         3335 &                   0 \\
mi-f                     &         ex-AE &     24945 &          40 &       2048 &   (8.21\%) &          341 &                   5 \\
mi-v                     &         ex-AE &     24945 &          40 &       3942 &   (15.8\%) &          341 &                   5 \\
mnist                    &          ODDS &      7603 &          78 &        700 &   (9.21\%) &            0 &                  22 \\
musk                     &          ODDS &      3062 &         166 &         97 &   (3.17\%) &            0 &                   0 \\
nasa                     &         ex-AE &      4687 &          32 &        755 &  (16.11\%) &            0 &                   0 \\
optdigits                &          ODDS &      5198 &          62 &        132 &   (2.54\%) &           18 &                   2 \\
pageblocks               &          ELKI &      5393 &          10 &        510 &   (9.46\%) &            0 &                   0 \\
parkinson                &          ELKI &       195 &          22 &        147 &  (75.38\%) &            0 &                   0 \\
pen-global               &     Goldstein &       808 &          16 &         90 &  (11.14\%) &            0 &                   0 \\
pen-local                &     Goldstein &      6723 &          16 &         10 &   (0.15\%) &            0 &                   0 \\
pendigits                &          ODDS &      6870 &          16 &        156 &   (2.27\%) &            0 &                   0 \\
pima                     &          ODDS &       768 &           8 &        268 &   (34.9\%) &            0 &                   0 \\
satellite                &          ODDS &      6435 &          36 &       2036 &  (31.64\%) &            0 &                   0 \\
satimage-2               &          ODDS &      5801 &          36 &         69 &   (1.19\%) &            2 &                   0 \\
seismic-bumps            &          ODDS &      2578 &          21 &        170 &   (6.59\%) &            6 &                   3 \\
shuttle                  &          ODDS &     49097 &           9 &       3511 &   (7.15\%) &            0 &                   0 \\
skin                     &       ADBench &     51433 &           3 &      14654 &  (28.49\%) &       193624 &                   0 \\
smtp                     &          ODDS &     71230 &           3 &         21 &   (0.03\%) &        23926 &                   0 \\
spambase                 &          GAAL &      4206 &          57 &       1678 &   (39.9\%) &            0 &                   0 \\
speech                   &          ODDS &      3686 &         400 &         61 &   (1.65\%) &            0 &                   0 \\
stamps                   &          ELKI &       340 &           9 &         31 &   (9.12\%) &            0 &                   0 \\
thyroid                  &          ODDS &      3656 &           6 &         93 &   (2.54\%) &          116 &                   0 \\
vertebral                &          ODDS &       240 &           6 &         30 &   (12.5\%) &            0 &                   0 \\
vowels                   &          ODDS &      1452 &          12 &         46 &   (3.17\%) &            4 &                   0 \\
waveform                 &          GAAL &      3442 &          21 &         99 &   (2.88\%) &            0 &                   0 \\
wbc                      &          ODDS &       377 &          30 &         20 &   (5.31\%) &            1 &                   0 \\
wbc2                     &       ADBench &       223 &           9 &         10 &   (4.48\%) &            0 &                   0 \\
wilt                     &          ELKI &      4819 &           5 &        257 &   (5.33\%) &            0 &                   0 \\
wine                     &          ODDS &       129 &          13 &         10 &   (7.75\%) &            0 &                   0 \\
wpbc                     &       ADBench &       198 &          33 &         47 &  (23.74\%) &            0 &                   0 \\
yeast                    &       ADBench &      1453 &           8 &        481 &   (33.1\%) &           31 &                   0 \\
yeast6                   &          EOAD &      1453 &           8 &         35 &   (2.41\%) &           31 &                   0 \\
\bottomrule
\end{tabular}

%% file: Tables/nemenyi_summary_truncated.tex
\begin{tabular}{lccccccccccccccc|r}
\toprule
{} & \rot{CBLOF} & \rot{sb-DeepSVDD} & \rot{SO-GAAL} & \rot{ALAD} & \rot{COF} & \rot{DeepSVDD} & \rot{ODIN} & \rot{LOF} & \rot{ensemble-LOF} & \rot{LMDD} & \rot{LODA} & \rot{SOD} & \rot{PCA} & \rot{VAE} & \rot{beta-VAE} & \rot{\shortstack[l]{\textbf{Mean}\\\textbf{AUC}}} \\
\midrule
EIF          &          ++ &                ++ &            ++ &         ++ &        ++ &             ++ &         ++ &        ++ &                 ++ &         ++ &         ++ &        ++ &        ++ &           &                &                                             0.773 \\
kth-NN       &          ++ &                ++ &            ++ &         ++ &        ++ &             ++ &         ++ &        ++ &                 ++ &         ++ &         ++ &        ++ &        ++ &         + &             ++ &                                             0.769 \\
kNN          &          ++ &                ++ &            ++ &         ++ &        ++ &             ++ &         ++ &        ++ &                 ++ &         ++ &         ++ &        ++ &        ++ &           &                &                                             0.766 \\
IF           &          ++ &                ++ &            ++ &         ++ &        ++ &             ++ &         ++ &        ++ &                 ++ &            &          + &           &        ++ &           &                &                                             0.764 \\
MCD          &          ++ &                ++ &            ++ &         ++ &        ++ &             ++ &         ++ &        ++ &                    &            &            &           &           &           &                &                                             0.756 \\
INNE         &          ++ &                ++ &            ++ &         ++ &        ++ &             ++ &         ++ &        ++ &                    &            &            &           &           &           &                &                                             0.749 \\
KDE          &          ++ &                ++ &            ++ &         ++ &        ++ &             ++ &         ++ &        ++ &                 ++ &          + &         ++ &         + &        ++ &           &                &                                             0.748 \\
gen2out      &          ++ &                ++ &            ++ &         ++ &        ++ &             ++ &         ++ &        ++ &                    &            &            &           &           &           &                &                                             0.748 \\
ABOD         &          ++ &                ++ &            ++ &         ++ &        ++ &             ++ &         ++ &        ++ &                  + &            &            &           &           &           &                &                                             0.747 \\
u-CBLOF      &          ++ &                ++ &            ++ &         ++ &        ++ &             ++ &         ++ &           &                    &            &            &           &           &           &                &                                             0.741 \\
COPOD        &          ++ &                ++ &            ++ &         ++ &        ++ &             ++ &         ++ &        ++ &                    &            &            &           &           &           &                &                                             0.738 \\
HBOS         &          ++ &                ++ &            ++ &         ++ &        ++ &             ++ &         ++ &           &                    &            &            &           &           &           &                &                                             0.725 \\
OCSVM        &          ++ &                ++ &            ++ &         ++ &        ++ &             ++ &         ++ &           &                    &            &            &           &           &           &                &                                             0.723 \\
AE           &          ++ &                ++ &            ++ &         ++ &           &             ++ &          + &           &                    &            &            &           &           &           &                &                                             0.720 \\
LUNAR        &          ++ &                ++ &            ++ &         ++ &           &             ++ &          + &           &                    &            &            &           &           &           &                &                                             0.711 \\
beta-VAE     &          ++ &                ++ &             + &         ++ &           &             ++ &            &           &                    &            &            &           &           &           &                &                                             0.710 \\
VAE          &          ++ &                ++ &            ++ &         ++ &           &             ++ &            &           &                    &            &            &           &           &           &                &                                             0.710 \\
ECOD         &          ++ &                ++ &            ++ &         ++ &           &             ++ &          + &           &                    &            &            &           &           &           &                &                                             0.709 \\
PCA          &             &                ++ &               &            &           &                &            &           &                    &            &            &           &           &           &                &                                             0.692 \\
SOD          &          ++ &                ++ &               &            &           &                &            &           &                    &            &            &           &           &           &                &                                             0.687 \\
LODA         &           + &                ++ &               &            &           &                &            &           &                    &            &            &           &           &           &                &                                             0.685 \\
LMDD         &          ++ &                ++ &               &            &           &                &            &           &                    &            &            &           &           &           &                &                                             0.684 \\
GMM          &          ++ &                ++ &            ++ &         ++ &           &             ++ &          + &           &                    &            &            &           &           &           &                &                                             0.683 \\
ensemble-LOF &             &                ++ &               &            &           &                &            &           &                    &            &            &           &           &           &                &                                             0.661 \\
LOF          &             &                   &               &            &           &                &            &           &                    &            &            &           &           &           &                &                                             0.630 \\
ODIN         &             &                   &               &            &           &                &            &           &                    &            &            &           &           &           &                &                                             0.595 \\
DeepSVDD     &             &                   &               &            &           &                &            &           &                    &            &            &           &           &      -{}- &           -{}- &                                             0.593 \\
COF          &             &                   &               &            &           &                &            &           &                    &            &            &           &           &           &                &                                             0.592 \\
ALAD         &             &                   &               &            &           &                &            &           &                    &            &            &           &           &      -{}- &           -{}- &                                             0.569 \\
SO-GAAL      &             &                   &               &            &           &                &            &           &                    &            &            &           &           &      -{}- &              - &                                             0.564 \\
sb-DeepSVDD  &             &                   &               &            &           &                &            &           &               -{}- &       -{}- &       -{}- &      -{}- &      -{}- &      -{}- &           -{}- &                                             0.553 \\
CBLOF        &             &                   &               &            &           &                &            &           &                    &       -{}- &          - &      -{}- &           &      -{}- &           -{}- &                                             0.519 \\
\bottomrule
\end{tabular}

%% file: Tables/nemenyi_summary_local_truncated.tex
\begin{tabular}{lcccccccccccccccc|r}
\toprule
{} & \rot{SO-GAAL} & \rot{PCA} & \rot{ECOD} & \rot{beta-VAE} & \rot{ALAD} & \rot{LMDD} & \rot{COPOD} & \rot{AE} & \rot{LODA} & \rot{OCSVM} & \rot{sb-DeepSVDD} & \rot{DeepSVDD} & \rot{VAE} & \rot{HBOS} & \rot{CBLOF} & \rot{gen2out} & \rot{\shortstack[l]{\textbf{Mean}\\\textbf{AUC}}} \\
\midrule
kNN          &            ++ &        ++ &         ++ &             ++ &         ++ &         ++ &          ++ &       ++ &         ++ &          ++ &                ++ &             ++ &        ++ &         ++ &          ++ &            ++ &                                             0.740 \\
ABOD         &            ++ &        ++ &         ++ &             ++ &         ++ &         ++ &          ++ &       ++ &         ++ &          ++ &                ++ &             ++ &        ++ &         ++ &          ++ &            ++ &                                             0.733 \\
kth-NN       &            ++ &        ++ &         ++ &             ++ &         ++ &         ++ &          ++ &       ++ &         ++ &          ++ &                ++ &              + &           &            &             &               &                                             0.722 \\
ensemble-LOF &            ++ &        ++ &         ++ &             ++ &         ++ &         ++ &          ++ &       ++ &            &             &                   &                &           &            &             &               &                                             0.714 \\
LOF          &            ++ &        ++ &         ++ &             ++ &         ++ &         ++ &          ++ &       ++ &          + &           + &                 + &                &           &            &             &               &                                             0.713 \\
COF          &            ++ &        ++ &         ++ &             ++ &         ++ &         ++ &          ++ &       ++ &            &             &                   &                &           &            &             &               &                                             0.713 \\
SOD          &            ++ &        ++ &         ++ &             ++ &         ++ &         ++ &          ++ &       ++ &            &           + &                 + &                &           &            &             &               &                                             0.711 \\
GMM          &            ++ &        ++ &         ++ &             ++ &            &         ++ &             &          &            &             &                   &                &           &            &             &               &                                             0.701 \\
KDE          &            ++ &        ++ &         ++ &             ++ &         ++ &         ++ &          ++ &       ++ &            &             &                 + &                &           &            &             &               &                                             0.693 \\
MCD          &               &        ++ &          + &              + &            &            &             &          &            &             &                   &                &           &            &             &               &                                             0.687 \\
LUNAR        &            ++ &        ++ &         ++ &             ++ &            &         ++ &             &          &            &             &                   &                &           &            &             &               &                                             0.677 \\
ODIN         &               &         + &            &                &            &            &             &          &            &             &                   &                &           &            &             &               &                                             0.672 \\
u-CBLOF      &               &           &            &                &            &            &             &          &            &             &                   &                &           &            &             &               &                                             0.635 \\
INNE         &               &           &            &                &            &            &             &          &            &             &                   &                &           &            &             &               &                                             0.624 \\
EIF          &               &           &            &                &            &            &             &          &            &             &                   &                &           &            &             &               &                                             0.614 \\
IF           &               &           &            &                &            &            &             &          &            &             &                   &                &           &            &             &               &                                             0.607 \\
gen2out      &               &           &            &                &            &            &             &          &            &             &                   &                &           &            &             &               &                                             0.585 \\
CBLOF        &               &           &            &                &            &            &             &          &            &             &                   &                &           &            &             &               &                                             0.577 \\
HBOS         &               &           &            &                &            &            &             &          &            &             &                   &                &           &            &             &               &                                             0.575 \\
VAE          &               &           &            &                &            &            &             &          &            &             &                   &                &           &            &             &               &                                             0.563 \\
DeepSVDD     &               &           &            &                &            &            &             &          &            &             &                   &                &           &            &             &               &                                             0.557 \\
sb-DeepSVDD  &               &           &            &                &            &            &             &          &            &             &                   &                &           &            &             &               &                                             0.546 \\
OCSVM        &               &           &            &                &            &            &             &          &            &             &                   &                &           &            &             &               &                                             0.543 \\
LODA         &               &           &            &                &            &            &             &          &            &             &                   &                &           &            &             &               &                                             0.534 \\
AE           &               &           &            &                &            &            &             &          &            &             &                   &                &           &            &             &               &                                             0.528 \\
COPOD        &               &           &            &                &            &            &             &          &            &             &                   &                &           &            &             &               &                                             0.518 \\
LMDD         &               &           &            &                &            &            &             &          &            &             &                   &                &           &            &             &               &                                             0.516 \\
ALAD         &               &           &            &                &            &            &             &          &            &             &                   &                &           &            &             &               &                                             0.511 \\
beta-VAE     &               &           &            &                &            &            &             &          &            &             &                   &                &           &            &             &               &                                             0.499 \\
ECOD         &               &           &            &                &            &            &             &          &            &             &                   &                &           &            &             &               &                                             0.492 \\
PCA          &               &           &            &                &            &            &             &          &            &             &                   &                &           &            &             &               &                                             0.475 \\
SO-GAAL      &               &           &            &                &            &            &             &          &            &             &                   &                &           &            &             &               &                                             0.423 \\
\bottomrule
\end{tabular}

%% file: Tables/nemenyi_summary_global_truncated.tex
\begin{tabular}{lccccccccccccccc|r}
\toprule
{} & \rot{CBLOF} & \rot{COF} & \rot{sb-DeepSVDD} & \rot{ODIN} & \rot{ALAD} & \rot{LOF} & \rot{DeepSVDD} & \rot{SO-GAAL} & \rot{ensemble-LOF} & \rot{GMM} & \rot{SOD} & \rot{LUNAR} & \rot{LODA} & \rot{LMDD} & \rot{PCA} & \rot{\shortstack[l]{\textbf{Mean}\\\textbf{AUC}}} \\
\midrule
EIF          &          ++ &        ++ &                ++ &         ++ &         ++ &        ++ &             ++ &            ++ &                 ++ &        ++ &        ++ &          ++ &         ++ &          + &         + &                                             0.824 \\
IF           &          ++ &        ++ &                ++ &         ++ &         ++ &        ++ &             ++ &            ++ &                 ++ &         + &        ++ &          ++ &          + &            &           &                                             0.814 \\
COPOD        &          ++ &        ++ &                ++ &         ++ &         ++ &        ++ &             ++ &            ++ &                 ++ &         + &        ++ &          ++ &            &            &           &                                             0.809 \\
gen2out      &          ++ &        ++ &                ++ &         ++ &         ++ &        ++ &             ++ &            ++ &                 ++ &           &        ++ &             &            &            &           &                                             0.801 \\
INNE         &          ++ &        ++ &                ++ &         ++ &         ++ &        ++ &             ++ &            ++ &                 ++ &           &        ++ &             &            &            &           &                                             0.790 \\
kth-NN       &          ++ &        ++ &                ++ &         ++ &         ++ &        ++ &             ++ &            ++ &                 ++ &           &        ++ &             &            &            &           &                                             0.785 \\
AE           &          ++ &        ++ &                ++ &         ++ &         ++ &        ++ &             ++ &            ++ &                 ++ &           &           &             &            &            &           &                                             0.782 \\
OCSVM        &          ++ &        ++ &                ++ &         ++ &         ++ &        ++ &             ++ &            ++ &                 ++ &           &        ++ &             &            &            &           &                                             0.781 \\
ECOD         &          ++ &        ++ &                ++ &         ++ &         ++ &        ++ &             ++ &            ++ &                 ++ &           &        ++ &             &            &            &           &                                             0.780 \\
MCD          &          ++ &        ++ &                ++ &         ++ &         ++ &        ++ &             ++ &            ++ &                 ++ &           &        ++ &             &            &            &           &                                             0.779 \\
beta-VAE     &          ++ &        ++ &                ++ &         ++ &         ++ &        ++ &             ++ &            ++ &                  + &           &           &             &            &            &           &                                             0.779 \\
u-CBLOF      &          ++ &        ++ &                ++ &         ++ &         ++ &        ++ &             ++ &            ++ &                  + &           &           &             &            &            &           &                                             0.776 \\
HBOS         &          ++ &        ++ &                ++ &         ++ &         ++ &        ++ &             ++ &            ++ &                 ++ &           &        ++ &             &            &            &           &                                             0.774 \\
kNN          &          ++ &        ++ &                ++ &         ++ &         ++ &        ++ &             ++ &            ++ &                 ++ &           &        ++ &             &            &            &           &                                             0.774 \\
KDE          &          ++ &        ++ &                ++ &         ++ &         ++ &        ++ &             ++ &            ++ &                 ++ &           &        ++ &             &            &            &           &                                             0.766 \\
PCA          &          ++ &        ++ &                ++ &         ++ &         ++ &        ++ &             ++ &               &                    &           &           &             &            &            &           &                                             0.763 \\
VAE          &          ++ &        ++ &                ++ &         ++ &         ++ &        ++ &             ++ &             + &                    &           &           &             &            &            &           &                                             0.758 \\
ABOD         &          ++ &        ++ &                ++ &         ++ &         ++ &        ++ &             ++ &            ++ &                    &           &           &             &            &            &           &                                             0.752 \\
LMDD         &          ++ &        ++ &                ++ &         ++ &         ++ &        ++ &             ++ &               &                    &           &           &             &            &            &           &                                             0.738 \\
LODA         &          ++ &        ++ &                ++ &         ++ &            &           &              + &               &                    &           &           &             &            &            &           &                                             0.734 \\
LUNAR        &          ++ &        ++ &                ++ &         ++ &            &           &                &               &                    &           &           &             &            &            &           &                                             0.722 \\
SOD          &             &           &                   &            &            &           &                &               &                    &           &           &             &            &            &           &                                             0.679 \\
GMM          &          ++ &        ++ &                ++ &         ++ &            &           &                &               &                    &           &           &             &            &            &           &                                             0.677 \\
ensemble-LOF &             &           &                   &            &            &           &                &               &                    &           &           &             &            &            &           &                                             0.643 \\
SO-GAAL      &             &           &                   &            &            &           &                &               &                    &           &           &             &            &            &           &                                             0.609 \\
DeepSVDD     &             &           &                   &            &            &           &                &               &                    &           &           &             &          - &       -{}- &      -{}- &                                             0.605 \\
LOF          &             &           &                   &            &            &           &                &               &                    &           &           &             &            &       -{}- &      -{}- &                                             0.603 \\
ALAD         &             &           &                   &            &            &           &                &               &                    &           &           &             &            &       -{}- &      -{}- &                                             0.588 \\
ODIN         &             &           &                   &            &            &           &                &               &                    &      -{}- &           &        -{}- &       -{}- &       -{}- &      -{}- &                                             0.571 \\
sb-DeepSVDD  &             &           &                   &            &            &           &                &               &                    &      -{}- &           &        -{}- &       -{}- &       -{}- &      -{}- &                                             0.556 \\
COF          &             &           &                   &            &            &           &                &               &                    &      -{}- &           &        -{}- &       -{}- &       -{}- &      -{}- &                                             0.553 \\
CBLOF        &             &           &                   &            &            &           &                &               &                    &      -{}- &           &        -{}- &       -{}- &       -{}- &      -{}- &                                             0.500 \\
\bottomrule
\end{tabular}

%% file: Tables/AUC_all_datasets.tex
\begin{tabular}{lrrrrrrrrrrrrrrrrrrrrrrrrrrrrrrrrrrrrrrrrrrrrrrrrr}
\toprule
{} & \rot{annthyroid} & \rot{wbc2} & \rot{mnist} & \rot{cover} & \rot{donors} & \rot{mi-f} & \rot{cardio} & \rot{speech} & \rot{breastw} & \rot{arrhythmia} & \rot{magic.gamma} & \rot{hrss} & \rot{pageblocks} & \rot{http} & \rot{vertebral} & \rot{wine} & \rot{seismic-bumps} & \rot{mammography} & \rot{landsat} & \rot{thyroid} & \rot{wbc} & \rot{campaign} & \rot{pima} & \rot{mi-v} & \rot{glass} & \rot{fault} & \rot{parkinson} & \rot{vowels} & \rot{yeast6} & \rot{wilt} & \rot{pendigits} & \rot{hepatitis} & \rot{satellite} & \rot{waveform} & \rot{shuttle} & \rot{satimage-2} & \rot{pen-local} & \rot{optdigits} & \rot{nasa} & \rot{letter} & \rot{skin} & \rot{spambase} & \rot{ionosphere} & \rot{smtp} & \rot{internetads} & \rot{musk} & \rot{aloi} & \rot{pen-global} & \rot{stamps} \\
\midrule
GMM          &             0.87 &       0.45 &        0.67 &        0.90 &         0.37 &       0.48 &         0.66 &         0.57 &          0.92 &             0.34 &              0.80 &       0.62 &             0.81 &       1.00 &            0.66 &       0.14 &                0.67 &              0.80 &          0.59 &          0.91 &      0.43 &           0.77 &       0.59 &       0.60 &        0.74 &        0.65 &            0.58 &         0.96 &         0.46 &       0.78 &            0.76 &            0.33 &            0.67 &           0.69 &          0.90 &             0.75 &            0.95 &            0.52 &       0.62 &         0.90 &       0.68 &           0.58 &             0.75 &       0.96 &              0.61 &       0.76 &       0.53 &             0.95 &         0.78 \\
CBLOF        &             0.37 &       0.20 &        0.64 &        0.87 &         0.46 &       0.57 &         0.53 &         0.52 &          0.48 &             0.48 &              0.50 &       0.54 &             0.64 &       0.31 &            0.47 &       0.30 &                0.45 &              0.43 &          0.55 &          0.24 &      0.31 &           0.41 &       0.42 &       0.55 &        0.52 &        0.59 &            0.43 &         0.83 &         0.40 &       0.57 &            0.45 &            0.37 &            0.63 &           0.69 &          0.98 &             0.32 &            0.66 &            0.47 &       0.45 &         0.58 &       0.62 &           0.59 &             0.72 &       0.65 &              0.55 &       0.61 &       0.54 &             0.61 &         0.34 \\
SO-GAAL      &             0.89 &       0.81 &        0.58 &        0.60 &         0.18 &       0.65 &         0.85 &         0.47 &          0.15 &             0.70 &              0.55 &       0.55 &             0.86 &       0.55 &            0.42 &       0.22 &                0.57 &              0.77 &          0.46 &          0.93 &      0.42 &           0.54 &       0.32 &       0.60 &        0.71 &        0.41 &            0.54 &         0.08 &         0.20 &       0.41 &            0.90 &            0.76 &            0.55 &           0.41 &          0.58 &             0.72 &            0.05 &            0.61 &       0.51 &         0.42 &       0.73 &           0.44 &             0.81 &       0.66 &              0.45 &       0.90 &       0.50 &             0.76 &         0.90 \\
INNE         &             0.92 &       0.92 &        0.82 &        0.95 &         0.32 &       0.56 &         0.87 &         0.47 &          0.82 &             0.73 &              0.71 &       0.55 &             0.96 &       1.00 &            0.60 &       0.81 &                0.71 &              0.81 &          0.54 &          0.99 &      0.93 &           0.81 &       0.66 &       0.65 &        0.75 &        0.52 &            0.46 &         0.88 &         0.71 &       0.56 &            0.87 &            0.69 &            0.75 &           0.74 &          0.98 &             1.00 &            0.74 &            0.54 &       0.59 &         0.69 &       0.73 &           0.59 &             0.90 &       0.96 &              0.69 &       0.99 &       0.53 &             0.90 &         0.83 \\
EIF          &             0.89 &       1.00 &        0.81 &        0.89 &         0.38 &       0.81 &         0.93 &         0.47 &          0.96 &             0.82 &              0.72 &       0.58 &             0.91 &       0.99 &            0.65 &       0.85 &                0.71 &              0.81 &          0.50 &          0.98 &      0.95 &           0.76 &       0.69 &       0.78 &        0.70 &        0.52 &            0.52 &         0.82 &         0.74 &       0.49 &            0.95 &            0.75 &            0.72 &           0.73 &          1.00 &             1.00 &            0.79 &            0.66 &       0.59 &         0.65 &       0.82 &           0.67 &             0.88 &       0.95 &              0.69 &       1.00 &       0.53 &             0.94 &         0.89 \\
sb-DeepSVDD  &             0.50 &       0.91 &        0.60 &        0.46 &         0.50 &       0.33 &         0.73 &         0.50 &          0.58 &             0.63 &              0.45 &       0.49 &             0.58 &       0.59 &            0.52 &       0.58 &                0.55 &              0.45 &          0.54 &          0.71 &      0.82 &           0.61 &       0.52 &       0.37 &        0.57 &        0.55 &            0.48 &         0.53 &         0.49 &       0.45 &            0.47 &            0.63 &            0.52 &           0.47 &          0.46 &             0.60 &            0.63 &            0.45 &       0.52 &         0.52 &       0.52 &           0.41 &             0.75 &       0.57 &              0.63 &       0.64 &       0.51 &             0.59 &         0.62 \\
ALAD         &             0.79 &       0.75 &        0.51 &        0.42 &         0.49 &       0.73 &         0.71 &         0.49 &          0.75 &             0.67 &              0.58 &       0.52 &             0.62 &       0.35 &            0.51 &       0.60 &                0.63 &              0.72 &          0.43 &          0.24 &      0.68 &           0.62 &       0.54 &       0.52 &        0.72 &        0.43 &            0.51 &         0.52 &         0.69 &       0.55 &            0.61 &            0.59 &            0.54 &           0.54 &          0.58 &             0.62 &            0.33 &            0.56 &       0.49 &         0.52 &       0.63 &           0.56 &             0.65 &       0.73 &              0.61 &       0.56 &       0.49 &             0.37 &         0.61 \\
AE           &             0.64 &       0.98 &        0.85 &        0.92 &         0.36 &       0.84 &         0.93 &         0.47 &          0.89 &             0.77 &              0.69 &       0.56 &             0.92 &       1.00 &            0.62 &       0.74 &                0.67 &              0.81 &          0.37 &          0.93 &      0.94 &           0.73 &       0.66 &       0.75 &        0.63 &        0.51 &            0.35 &         0.79 &         0.69 &       0.33 &            0.93 &            0.73 &            0.60 &           0.64 &          0.99 &             0.98 &            0.55 &            0.51 &       0.50 &         0.56 &       0.71 &           0.55 &             0.82 &       0.89 &              0.61 &       1.00 &       0.55 &             0.89 &         0.88 \\
MCD          &             0.93 &       0.99 &        0.83 &        0.84 &         0.37 &       0.40 &         0.81 &         0.50 &          0.94 &             0.72 &              0.74 &       0.59 &             0.93 &       1.00 &            0.61 &       0.77 &                0.73 &              0.76 &          0.56 &          0.99 &      0.94 &           0.78 &       0.68 &       0.58 &        0.77 &        0.50 &            0.62 &         0.92 &         0.67 &       0.85 &            0.84 &            0.76 &            0.76 &           0.59 &          1.00 &             0.99 &            0.74 &            0.41 &       0.61 &         0.81 &       0.82 &           0.49 &             0.95 &       0.96 &              0.75 &       1.00 &       0.53 &             0.91 &         0.84 \\
kth-NN       &             0.92 &       0.99 &        0.82 &        0.79 &         0.48 &       0.41 &         0.84 &         0.48 &          0.95 &             0.80 &              0.79 &       0.57 &             0.95 &       1.00 &            0.66 &       0.90 &                0.73 &              0.79 &          0.61 &          0.99 &      0.95 &           0.81 &       0.71 &       0.59 &        0.79 &        0.64 &            0.60 &         0.95 &         0.68 &       0.69 &            0.82 &            0.81 &            0.71 &           0.74 &          0.84 &             0.97 &            0.98 &            0.48 &       0.65 &         0.81 &       0.70 &           0.67 &             0.90 &       0.96 &              0.71 &       0.63 &       0.58 &             0.98 &         0.89 \\
COPOD        &             0.77 &       0.99 &        0.77 &        0.88 &         0.36 &       0.70 &         0.93 &         0.49 &          0.98 &             0.80 &              0.68 &       0.60 &             0.88 &       0.99 &            0.67 &       0.87 &                0.71 &              0.87 &          0.42 &          0.94 &      0.97 &           0.78 &       0.65 &       0.67 &        0.65 &        0.46 &            0.54 &         0.51 &         0.81 &       0.34 &            0.90 &            0.80 &            0.63 &           0.73 &          0.99 &             0.97 &            0.52 &            0.68 &       0.54 &         0.56 &       0.86 &           0.68 &             0.80 &       0.93 &              0.68 &       0.95 &       0.52 &             0.79 &         0.93 \\
LOF          &             0.49 &       0.75 &        0.60 &        0.51 &         0.50 &       0.57 &         0.53 &         0.53 &          0.40 &             0.61 &              0.70 &       0.56 &             0.67 &       0.69 &            0.53 &       0.76 &                0.55 &              0.68 &          0.54 &          0.60 &      0.92 &           0.56 &       0.61 &       0.55 &        0.79 &        0.57 &            0.54 &         0.93 &         0.47 &       0.65 &            0.51 &            0.72 &            0.54 &           0.73 &          0.58 &             0.55 &            0.98 &            0.50 &       0.56 &         0.87 &       0.56 &           0.50 &             0.90 &       0.72 &              0.64 &       0.55 &       0.74 &             0.72 &         0.64 \\
ODIN         &             0.50 &       0.80 &        0.62 &        0.52 &         0.50 &       0.49 &         0.57 &         0.65 &          0.51 &             0.63 &              0.65 &       0.58 &             0.58 &       0.53 &            0.50 &       0.66 &                0.55 &              0.61 &          0.49 &          0.57 &      0.79 &           0.56 &       0.56 &       0.53 &        0.64 &        0.59 &            0.43 &         0.87 &         0.44 &       0.66 &            0.51 &            0.67 &            0.49 &           0.66 &          0.50 &             0.55 &            0.88 &            0.51 &       0.56 &         0.86 &       0.51 &           0.53 &             0.85 &       0.45 &              0.57 &       0.55 &       0.74 &             0.66 &         0.59 \\
ensemble-LOF &             0.52 &       0.92 &        0.63 &        0.50 &         0.48 &       0.62 &         0.60 &         0.55 &          0.39 &             0.62 &              0.70 &       0.55 &             0.71 &       0.91 &            0.55 &       0.88 &                0.52 &              0.69 &          0.55 &          0.74 &      0.95 &           0.46 &       0.61 &       0.60 &        0.79 &        0.56 &            0.54 &         0.94 &         0.47 &       0.64 &            0.53 &            0.64 &            0.58 &           0.72 &          0.62 &             0.66 &            0.99 &            0.52 &       0.55 &         0.87 &       0.59 &           0.51 &             0.91 &       0.91 &              0.68 &       0.63 &       0.75 &             0.81 &         0.70 \\
beta-VAE     &             0.67 &       0.99 &        0.85 &        0.93 &         0.40 &       0.83 &         0.95 &         0.47 &          0.73 &             0.77 &              0.67 &       0.55 &             0.91 &       1.00 &            0.62 &       0.82 &                0.66 &              0.83 &          0.39 &          0.95 &      0.94 &           0.73 &       0.66 &       0.75 &        0.59 &        0.49 &            0.35 &         0.64 &         0.77 &       0.33 &            0.94 &            0.75 &            0.60 &           0.65 &          0.99 &             0.98 &            0.44 &            0.51 &       0.49 &         0.52 &       0.71 &           0.55 &             0.80 &       0.89 &              0.61 &       0.86 &       0.55 &             0.81 &         0.91 \\
COF          &             0.48 &       0.60 &        0.61 &        0.51 &         0.50 &       0.54 &         0.51 &         0.56 &          0.43 &             0.65 &              0.65 &       0.52 &             0.58 &       0.65 &            0.52 &       0.40 &                0.55 &              0.63 &          0.53 &          0.50 &      0.86 &           0.50 &       0.60 &       0.53 &        0.79 &        0.56 &            0.61 &         0.94 &         0.39 &       0.62 &            0.51 &            0.58 &            0.52 &           0.71 &          0.56 &             0.56 &            0.94 &            0.50 &       0.54 &         0.86 &       0.54 &           0.47 &             0.89 &       0.47 &              0.65 &       0.50 &       0.77 &             0.64 &         0.51 \\
DeepSVDD     &             0.55 &       0.94 &        0.63 &        0.47 &         0.45 &       0.42 &         0.69 &         0.51 &          0.64 &             0.67 &              0.50 &       0.50 &             0.60 &       0.61 &            0.57 &       0.71 &                0.59 &              0.48 &          0.60 &          0.66 &      0.85 &           0.64 &       0.52 &       0.48 &        0.46 &        0.52 &            0.51 &         0.60 &         0.64 &       0.49 &            0.58 &            0.66 &            0.59 &           0.45 &          0.49 &             0.75 &            0.72 &            0.54 &       0.46 &         0.62 &       0.43 &           0.47 &             0.77 &       0.87 &              0.68 &       0.73 &       0.50 &             0.67 &         0.62 \\
u-CBLOF      &             0.91 &       0.99 &        0.82 &        0.89 &         0.39 &       0.62 &         0.85 &         0.47 &          0.92 &             0.79 &              0.70 &       0.56 &             0.92 &       1.00 &            0.58 &       0.64 &                0.69 &              0.74 &          0.56 &          0.99 &      0.93 &           0.80 &       0.66 &       0.59 &        0.79 &        0.56 &            0.58 &         0.86 &         0.65 &       0.53 &            0.91 &            0.73 &            0.76 &           0.71 &          0.99 &             1.00 &            0.81 &            0.51 &       0.44 &         0.70 &       0.70 &           0.51 &             0.87 &       0.93 &              0.70 &       0.86 &       0.54 &             0.90 &         0.78 \\
OCSVM        &             0.94 &       0.98 &        0.70 &        0.93 &         0.42 &       0.51 &         0.88 &         0.47 &          0.83 &             0.80 &              0.67 &       0.56 &             0.93 &       1.00 &            0.62 &       0.86 &                0.71 &              0.82 &          0.38 &          0.99 &      0.95 &           0.80 &       0.63 &       0.62 &        0.65 &        0.47 &            0.45 &         0.72 &         0.73 &       0.44 &            0.93 &            0.76 &            0.63 &           0.65 &          0.98 &             0.99 &            0.47 &            0.51 &       0.52 &         0.61 &       0.74 &           0.52 &             0.84 &       0.96 &              0.69 &       0.87 &       0.54 &             0.87 &         0.88 \\
ABOD         &             0.91 &       0.99 &        0.81 &        0.80 &         0.50 &       0.40 &         0.76 &         0.57 &          0.95 &             0.82 &              0.80 &       0.52 &             0.94 &       0.99 &            0.64 &       0.76 &                0.74 &              0.79 &          0.58 &          0.98 &      0.94 &           0.79 &       0.70 &       0.59 &        0.80 &        0.68 &            0.63 &         0.96 &         0.62 &       0.69 &            0.77 &            0.79 &            0.66 &           0.70 &          0.79 &             0.95 &            0.96 &            0.47 &       0.63 &         0.82 &       0.69 &           0.52 &             0.93 &       0.95 &              0.74 &       0.18 &       0.61 &             0.98 &         0.85 \\
PCA          &             0.69 &       0.99 &        0.85 &        0.93 &         0.33 &       0.84 &         0.95 &         0.47 &          0.51 &             0.78 &              0.65 &       0.55 &             0.90 &       1.00 &            0.51 &       0.81 &                0.64 &              0.75 &          0.40 &          0.96 &      0.93 &           0.74 &       0.59 &       0.74 &        0.53 &        0.50 &            0.31 &         0.65 &         0.71 &       0.31 &            0.92 &            0.76 &            0.63 &           0.60 &          0.99 &             0.97 &            0.34 &            0.48 &       0.46 &         0.52 &       0.64 &           0.55 &             0.80 &       0.94 &              0.61 &       1.00 &       0.55 &             0.78 &         0.85 \\
kNN          &             0.91 &       0.99 &        0.81 &        0.79 &         0.49 &       0.41 &         0.81 &         0.51 &          0.95 &             0.80 &              0.80 &       0.57 &             0.95 &       1.00 &            0.65 &       0.81 &                0.73 &              0.79 &          0.60 &          0.99 &      0.95 &           0.80 &       0.71 &       0.58 &        0.82 &        0.66 &            0.64 &         0.97 &         0.64 &       0.69 &            0.79 &            0.79 &            0.70 &           0.74 &          0.82 &             0.95 &            0.98 &            0.48 &       0.65 &         0.85 &       0.70 &           0.65 &             0.92 &       0.96 &              0.73 &       0.54 &       0.60 &             0.98 &         0.86 \\
VAE          &             0.65 &       0.99 &        0.84 &        0.94 &         0.37 &       0.64 &         0.87 &         0.47 &          0.78 &             0.77 &              0.69 &       0.55 &             0.91 &       1.00 &            0.62 &       0.79 &                0.67 &              0.80 &          0.51 &          0.94 &      0.90 &           0.72 &       0.66 &       0.65 &        0.62 &        0.60 &            0.45 &         0.69 &         0.77 &       0.33 &            0.92 &            0.72 &            0.63 &           0.66 &          0.79 &             0.87 &            0.63 &            0.53 &       0.54 &         0.63 &       0.71 &           0.56 &             0.84 &       0.90 &              0.61 &       0.80 &       0.55 &             0.86 &         0.86 \\
LODA         &             0.72 &       0.98 &        0.82 &        0.91 &         0.94 &       0.84 &         0.76 &         0.55 &          0.92 &             0.51 &              0.69 &       0.49 &             0.78 &       0.98 &            0.65 &       0.62 &                0.57 &              0.83 &          0.39 &          0.85 &      0.91 &           0.51 &       0.56 &       0.69 &        0.18 &        0.40 &            0.75 &         0.58 &         0.78 &       0.37 &            0.91 &            0.61 &            0.62 &           0.70 &          0.67 &             0.98 &            0.64 &            0.53 &       0.57 &         0.55 &       0.76 &           0.59 &             0.53 &       0.83 &              0.60 &       0.99 &       0.52 &             0.58 &         0.85 \\
IF           &             0.82 &       1.00 &        0.81 &        0.89 &         0.37 &       0.79 &         0.93 &         0.47 &          0.96 &             0.81 &              0.73 &       0.58 &             0.90 &       0.99 &            0.64 &       0.79 &                0.67 &              0.82 &          0.48 &          0.98 &      0.95 &           0.72 &       0.67 &       0.77 &        0.70 &        0.58 &            0.49 &         0.78 &         0.74 &       0.46 &            0.95 &            0.69 &            0.70 &           0.73 &          1.00 &             0.99 &            0.80 &            0.73 &       0.57 &         0.64 &       0.82 &           0.62 &             0.86 &       0.94 &              0.68 &       1.00 &       0.54 &             0.93 &         0.89 \\
ECOD         &             0.78 &       0.99 &        0.75 &        0.93 &         0.41 &       0.60 &         0.95 &         0.49 &          0.82 &             0.81 &              0.64 &       0.59 &             0.91 &       0.99 &            0.58 &       0.71 &                0.68 &              0.78 &          0.37 &          0.98 &      0.91 &           0.77 &       0.52 &       0.65 &        0.68 &        0.47 &            0.41 &         0.41 &         0.71 &       0.42 &            0.91 &            0.79 &            0.75 &           0.72 &          1.00 &             0.97 &            0.45 &            0.60 &       0.44 &         0.53 &       0.72 &           0.64 &             0.77 &       0.91 &              0.68 &       0.96 &       0.53 &             0.79 &         0.92 \\
LMDD         &             0.92 &       0.99 &        0.75 &        0.91 &         0.64 &       0.59 &         0.88 &         0.48 &          0.53 &             0.80 &              0.63 &       0.50 &             0.78 &       1.00 &            0.66 &       0.78 &                0.67 &              0.78 &          0.45 &          0.99 &      0.96 &           0.70 &       0.62 &       0.62 &        0.58 &        0.40 &            0.47 &         0.61 &         0.70 &       0.36 &            0.94 &            0.83 &            0.42 &           0.64 &          0.99 &             0.48 &            0.45 &            0.50 &       0.49 &         0.52 &       0.51 &           0.50 &             0.74 &       0.94 &              0.68 &       0.97 &       0.50 &             0.78 &         0.90 \\
gen2out      &             0.86 &       1.00 &        0.78 &        0.92 &         0.41 &       0.70 &         0.95 &         0.46 &          0.97 &             0.78 &              0.72 &       0.58 &             0.91 &       1.00 &            0.63 &       0.87 &                0.71 &              0.81 &          0.48 &          0.98 &      0.95 &           0.66 &       0.66 &       0.70 &        0.74 &        0.50 &            0.52 &         0.75 &         0.68 &       0.47 &            0.95 &            0.72 &            0.72 &           0.63 &          0.99 &             0.99 &            0.66 &            0.57 &       0.53 &         0.68 &       0.81 &           0.66 &             0.76 &       0.96 &              0.56 &       1.00 &       0.51 &             0.86 &         0.91 \\
LUNAR        &             0.69 &       0.97 &        0.76 &        0.73 &         0.46 &       0.42 &         0.66 &         0.46 &          0.95 &             0.79 &              0.81 &       0.60 &             0.72 &       0.99 &            0.63 &       0.69 &                0.73 &              0.79 &          0.59 &          0.92 &      0.92 &           0.67 &       0.69 &       0.62 &        0.80 &        0.70 &            0.47 &         0.83 &         0.64 &       0.48 &            0.72 &            0.66 &            0.66 &           0.74 &          0.65 &             0.89 &            0.90 &            0.43 &       0.54 &         0.74 &       0.67 &           0.51 &             0.91 &       0.95 &              0.65 &       0.73 &       0.71 &             0.81 &         0.80 \\
KDE          &             0.94 &       0.97 &        0.73 &        0.92 &         0.39 &       0.37 &         0.81 &         0.43 &          0.96 &             0.67 &              0.68 &       0.55 &             0.94 &       1.00 &            0.67 &       0.77 &                0.74 &              0.81 &          0.60 &          0.98 &      0.94 &           0.84 &       0.70 &       0.59 &        0.77 &        0.63 &            0.62 &         0.89 &         0.73 &       0.50 &            0.96 &            0.78 &            0.78 &           0.76 &          0.93 &             0.99 &            0.87 &            0.51 &       0.64 &         0.88 &       0.75 &           0.60 &             0.94 &       0.97 &              0.66 &       0.08 &       0.59 &             0.96 &         0.87 \\
HBOS         &             0.68 &       0.99 &        0.35 &        0.64 &         0.42 &       0.43 &         0.80 &         0.46 &          0.97 &             0.80 &              0.71 &       0.57 &             0.75 &       0.98 &            0.64 &       0.91 &                0.71 &              0.80 &          0.58 &          0.97 &      0.96 &           0.78 &       0.70 &       0.62 &        0.68 &        0.59 &            0.52 &         0.67 &         0.74 &       0.36 &            0.93 &            0.78 &            0.76 &           0.68 &          0.99 &             0.98 &            0.73 &            0.87 &       0.49 &         0.60 &       0.85 &           0.64 &             0.76 &       0.83 &              0.68 &       1.00 &       0.50 &             0.77 &         0.91 \\
SOD          &             0.78 &       0.96 &        0.71 &        0.63 &         0.43 &       0.39 &         0.66 &         0.56 &          0.88 &             0.76 &              0.75 &       0.55 &             0.76 &       0.88 &            0.59 &       0.47 &                0.71 &              0.74 &          0.57 &          0.94 &      0.95 &           0.73 &       0.59 &       0.56 &        0.72 &        0.64 &            0.70 &         0.92 &         0.58 &       0.59 &            0.66 &            0.57 &            0.58 &           0.62 &          0.74 &             0.78 &            0.92 &            0.55 &       0.59 &         0.89 &       0.59 &           0.54 &             0.90 &       0.82 &              0.53 &       0.45 &       0.66 &             0.81 &         0.74 \\
\bottomrule
\end{tabular}

%% file: Tables/nemenyi_table_all_datasets.tex
\begin{tabular}{lllllllllllllllllllllllllllllllll}
\toprule
{} & {GMM} & {CBLOF} & {SO-GAAL} & {INNE} & {EIF} & {sb-DeepSVDD} & {ALAD} & {AE} & {MCD} & {kth-NN} & {COPOD} & {LOF} & {ODIN} & {ensemble-LOF} & {beta-VAE} & {COF} & {DeepSVDD} & {u-CBLOF} & {OCSVM} & {ABOD} & {PCA} & {kNN} & {VAE} & {LODA} & {IF} & {ECOD} & {LMDD} & {gen2out} & {LUNAR} & {KDE} & {HBOS} & {SOD} \\
\midrule
GMM & 1.0 & \textbf{0.001} & \textbf{0.016} & 0.9 & 0.272 & \textbf{0.001} & \textbf{0.002} & 0.9 & 0.9 & 0.138 & 0.9 & 0.878 & 0.072 & 0.9 & 0.9 & 0.147 & \textbf{0.007} & 0.9 & 0.9 & 0.9 & 0.9 & 0.272 & 0.9 & 0.9 & 0.827 & 0.9 & 0.9 & 0.9 & 0.9 & 0.666 & 0.9 & 0.9 \\
CBLOF & \textbf{0.001} & 1.0 & 0.9 & \textbf{0.001} & \textbf{0.001} & 0.9 & 0.9 & \textbf{0.001} & \textbf{0.001} & \textbf{0.001} & \textbf{0.001} & 0.725 & 0.9 & 0.125 & \textbf{0.002} & 0.9 & 0.9 & \textbf{0.001} & \textbf{0.001} & \textbf{0.001} & 0.115 & \textbf{0.001} & \textbf{0.001} & 0.088 & \textbf{0.001} & \textbf{0.001} & \textbf{0.048} & \textbf{0.001} & \textbf{0.001} & \textbf{0.001} & \textbf{0.001} & \textbf{0.045} \\
SO-GAAL & \textbf{0.016} & 0.9 & 1.0 & \textbf{0.001} & \textbf{0.001} & 0.9 & 0.9 & \textbf{0.021} & \textbf{0.001} & \textbf{0.001} & \textbf{0.001} & 0.9 & 0.9 & 0.64 & 0.058 & 0.9 & 0.9 & \textbf{0.001} & \textbf{0.001} & \textbf{0.001} & 0.622 & \textbf{0.001} & \textbf{0.034} & 0.563 & \textbf{0.001} & \textbf{0.019} & 0.433 & \textbf{0.001} & \textbf{0.021} & \textbf{0.001} & \textbf{0.001} & 0.416 \\
INNE & 0.9 & \textbf{0.001} & \textbf{0.001} & 1.0 & 0.9 & \textbf{0.001} & \textbf{0.001} & 0.9 & 0.9 & 0.9 & 0.9 & \textbf{0.021} & \textbf{0.001} & 0.35 & 0.9 & \textbf{0.001} & \textbf{0.001} & 0.9 & 0.9 & 0.9 & 0.372 & 0.9 & 0.9 & 0.442 & 0.9 & 0.9 & 0.571 & 0.9 & 0.9 & 0.9 & 0.9 & 0.585 \\
EIF & 0.272 & \textbf{0.001} & \textbf{0.001} & 0.9 & 1.0 & \textbf{0.001} & \textbf{0.001} & 0.226 & 0.9 & 0.9 & 0.9 & \textbf{0.001} & \textbf{0.001} & \textbf{0.002} & 0.104 & \textbf{0.001} & \textbf{0.001} & 0.9 & 0.754 & 0.9 & \textbf{0.002} & 0.9 & 0.161 & \textbf{0.003} & 0.9 & 0.241 & \textbf{0.007} & 0.9 & 0.226 & 0.9 & 0.857 & \textbf{0.007} \\
sb-DeepSVDD & \textbf{0.001} & 0.9 & 0.9 & \textbf{0.001} & \textbf{0.001} & 1.0 & 0.9 & \textbf{0.001} & \textbf{0.001} & \textbf{0.001} & \textbf{0.001} & 0.265 & 0.9 & \textbf{0.013} & \textbf{0.001} & 0.9 & 0.9 & \textbf{0.001} & \textbf{0.001} & \textbf{0.001} & \textbf{0.012} & \textbf{0.001} & \textbf{0.001} & \textbf{0.008} & \textbf{0.001} & \textbf{0.001} & \textbf{0.004} & \textbf{0.001} & \textbf{0.001} & \textbf{0.001} & \textbf{0.001} & \textbf{0.003} \\
ALAD & \textbf{0.002} & 0.9 & 0.9 & \textbf{0.001} & \textbf{0.001} & 0.9 & 1.0 & \textbf{0.002} & \textbf{0.001} & \textbf{0.001} & \textbf{0.001} & 0.9 & 0.9 & 0.268 & \textbf{0.008} & 0.9 & 0.9 & \textbf{0.001} & \textbf{0.001} & \textbf{0.001} & 0.251 & \textbf{0.001} & \textbf{0.004} & 0.203 & \textbf{0.001} & \textbf{0.002} & 0.123 & \textbf{0.001} & \textbf{0.002} & \textbf{0.001} & \textbf{0.001} & 0.115 \\
AE & 0.9 & \textbf{0.001} & \textbf{0.021} & 0.9 & 0.226 & \textbf{0.001} & \textbf{0.002} & 1.0 & 0.9 & 0.112 & 0.9 & 0.9 & 0.091 & 0.9 & 0.9 & 0.181 & \textbf{0.01} & 0.9 & 0.9 & 0.9 & 0.9 & 0.226 & 0.9 & 0.9 & 0.776 & 0.9 & 0.9 & 0.9 & 0.9 & 0.615 & 0.9 & 0.9 \\
MCD & 0.9 & \textbf{0.001} & \textbf{0.001} & 0.9 & 0.9 & \textbf{0.001} & \textbf{0.001} & 0.9 & 1.0 & 0.9 & 0.9 & \textbf{0.012} & \textbf{0.001} & 0.254 & 0.9 & \textbf{0.001} & \textbf{0.001} & 0.9 & 0.9 & 0.9 & 0.272 & 0.9 & 0.9 & 0.329 & 0.9 & 0.9 & 0.474 & 0.9 & 0.9 & 0.9 & 0.9 & 0.49 \\
kth-NN & 0.138 & \textbf{0.001} & \textbf{0.001} & 0.9 & 0.9 & \textbf{0.001} & \textbf{0.001} & 0.112 & 0.9 & 1.0 & 0.9 & \textbf{0.001} & \textbf{0.001} & \textbf{0.001} & \textbf{0.045} & \textbf{0.001} & \textbf{0.001} & 0.776 & 0.578 & 0.9 & \textbf{0.001} & 0.9 & 0.075 & \textbf{0.001} & 0.9 & 0.121 & \textbf{0.002} & 0.9 & 0.112 & 0.9 & 0.681 & \textbf{0.002} \\
COPOD & 0.9 & \textbf{0.001} & \textbf{0.001} & 0.9 & 0.9 & \textbf{0.001} & \textbf{0.001} & 0.9 & 0.9 & 0.9 & 1.0 & \textbf{0.042} & \textbf{0.001} & 0.494 & 0.9 & \textbf{0.001} & \textbf{0.001} & 0.9 & 0.9 & 0.9 & 0.512 & 0.9 & 0.9 & 0.571 & 0.9 & 0.9 & 0.695 & 0.9 & 0.9 & 0.9 & 0.9 & 0.71 \\
LOF & 0.878 & 0.725 & 0.9 & \textbf{0.021} & \textbf{0.001} & 0.265 & 0.9 & 0.9 & \textbf{0.012} & \textbf{0.001} & \textbf{0.042} & 1.0 & 0.9 & 0.9 & 0.9 & 0.9 & 0.9 & 0.208 & 0.407 & \textbf{0.003} & 0.9 & \textbf{0.001} & 0.9 & 0.9 & \textbf{0.001} & 0.9 & 0.9 & \textbf{0.017} & 0.9 & \textbf{0.001} & 0.293 & 0.9 \\
ODIN & 0.072 & 0.9 & 0.9 & \textbf{0.001} & \textbf{0.001} & 0.9 & 0.9 & 0.091 & \textbf{0.001} & \textbf{0.001} & \textbf{0.001} & 0.9 & 1.0 & 0.9 & 0.203 & 0.9 & 0.9 & \textbf{0.002} & \textbf{0.006} & \textbf{0.001} & 0.9 & \textbf{0.001} & 0.133 & 0.849 & \textbf{0.001} & 0.085 & 0.725 & \textbf{0.001} & 0.091 & \textbf{0.001} & \textbf{0.003} & 0.71 \\
ensemble-LOF & 0.9 & 0.125 & 0.64 & 0.35 & \textbf{0.002} & \textbf{0.013} & 0.268 & 0.9 & 0.254 & \textbf{0.001} & 0.494 & 0.9 & 0.9 & 1.0 & 0.9 & 0.9 & 0.501 & 0.853 & 0.9 & 0.096 & 0.9 & \textbf{0.002} & 0.9 & 0.9 & \textbf{0.044} & 0.9 & 0.9 & 0.304 & 0.9 & \textbf{0.018} & 0.9 & 0.9 \\
beta-VAE & 0.9 & \textbf{0.002} & 0.058 & 0.9 & 0.104 & \textbf{0.001} & \textbf{0.008} & 0.9 & 0.9 & \textbf{0.045} & 0.9 & 0.9 & 0.203 & 0.9 & 1.0 & 0.354 & \textbf{0.028} & 0.9 & 0.9 & 0.747 & 0.9 & 0.104 & 0.9 & 0.9 & 0.585 & 0.9 & 0.9 & 0.9 & 0.9 & 0.416 & 0.9 & 0.9 \\
COF & 0.147 & 0.9 & 0.9 & \textbf{0.001} & \textbf{0.001} & 0.9 & 0.9 & 0.181 & \textbf{0.001} & \textbf{0.001} & \textbf{0.001} & 0.9 & 0.9 & 0.9 & 0.354 & 1.0 & 0.9 & \textbf{0.004} & \textbf{0.015} & \textbf{0.001} & 0.9 & \textbf{0.001} & 0.251 & 0.9 & \textbf{0.001} & 0.169 & 0.886 & \textbf{0.001} & 0.181 & \textbf{0.001} & \textbf{0.008} & 0.871 \\
DeepSVDD & \textbf{0.007} & 0.9 & 0.9 & \textbf{0.001} & \textbf{0.001} & 0.9 & 0.9 & \textbf{0.01} & \textbf{0.001} & \textbf{0.001} & \textbf{0.001} & 0.9 & 0.9 & 0.501 & \textbf{0.028} & 0.9 & 1.0 & \textbf{0.001} & \textbf{0.001} & \textbf{0.001} & 0.482 & \textbf{0.001} & \textbf{0.016} & 0.416 & \textbf{0.001} & \textbf{0.009} & 0.279 & \textbf{0.001} & \textbf{0.01} & \textbf{0.001} & \textbf{0.001} & 0.265 \\
u-CBLOF & 0.9 & \textbf{0.001} & \textbf{0.001} & 0.9 & 0.9 & \textbf{0.001} & \textbf{0.001} & 0.9 & 0.9 & 0.776 & 0.9 & 0.208 & \textbf{0.002} & 0.853 & 0.9 & \textbf{0.004} & \textbf{0.001} & 1.0 & 0.9 & 0.9 & 0.871 & 0.9 & 0.9 & 0.9 & 0.9 & 0.9 & 0.9 & 0.9 & 0.9 & 0.9 & 0.9 & 0.9 \\
OCSVM & 0.9 & \textbf{0.001} & \textbf{0.001} & 0.9 & 0.754 & \textbf{0.001} & \textbf{0.001} & 0.9 & 0.9 & 0.578 & 0.9 & 0.407 & \textbf{0.006} & 0.9 & 0.9 & \textbf{0.015} & \textbf{0.001} & 0.9 & 1.0 & 0.9 & 0.9 & 0.754 & 0.9 & 0.9 & 0.9 & 0.9 & 0.9 & 0.9 & 0.9 & 0.9 & 0.9 & 0.9 \\
ABOD & 0.9 & \textbf{0.001} & \textbf{0.001} & 0.9 & 0.9 & \textbf{0.001} & \textbf{0.001} & 0.9 & 0.9 & 0.9 & 0.9 & \textbf{0.003} & \textbf{0.001} & 0.096 & 0.747 & \textbf{0.001} & \textbf{0.001} & 0.9 & 0.9 & 1.0 & 0.104 & 0.9 & 0.849 & 0.133 & 0.9 & 0.9 & 0.22 & 0.9 & 0.9 & 0.9 & 0.9 & 0.232 \\
PCA & 0.9 & 0.115 & 0.622 & 0.372 & \textbf{0.002} & \textbf{0.012} & 0.251 & 0.9 & 0.272 & \textbf{0.001} & 0.512 & 0.9 & 0.9 & 0.9 & 0.9 & 0.9 & 0.482 & 0.871 & 0.9 & 0.104 & 1.0 & \textbf{0.002} & 0.9 & 0.9 & \textbf{0.048} & 0.9 & 0.9 & 0.322 & 0.9 & \textbf{0.02} & 0.9 & 0.9 \\
kNN & 0.272 & \textbf{0.001} & \textbf{0.001} & 0.9 & 0.9 & \textbf{0.001} & \textbf{0.001} & 0.226 & 0.9 & 0.9 & 0.9 & \textbf{0.001} & \textbf{0.001} & \textbf{0.002} & 0.104 & \textbf{0.001} & \textbf{0.001} & 0.9 & 0.754 & 0.9 & \textbf{0.002} & 1.0 & 0.161 & \textbf{0.003} & 0.9 & 0.241 & \textbf{0.007} & 0.9 & 0.226 & 0.9 & 0.857 & \textbf{0.007} \\
VAE & 0.9 & \textbf{0.001} & \textbf{0.034} & 0.9 & 0.161 & \textbf{0.001} & \textbf{0.004} & 0.9 & 0.9 & 0.075 & 0.9 & 0.9 & 0.133 & 0.9 & 0.9 & 0.251 & \textbf{0.016} & 0.9 & 0.9 & 0.849 & 0.9 & 0.161 & 1.0 & 0.9 & 0.688 & 0.9 & 0.9 & 0.9 & 0.9 & 0.527 & 0.9 & 0.9 \\
LODA & 0.9 & 0.088 & 0.563 & 0.442 & \textbf{0.003} & \textbf{0.008} & 0.203 & 0.9 & 0.329 & \textbf{0.001} & 0.571 & 0.9 & 0.849 & 0.9 & 0.9 & 0.9 & 0.416 & 0.9 & 0.9 & 0.133 & 0.9 & \textbf{0.003} & 0.9 & 1.0 & 0.065 & 0.9 & 0.9 & 0.389 & 0.9 & \textbf{0.028} & 0.9 & 0.9 \\
IF & 0.827 & \textbf{0.001} & \textbf{0.001} & 0.9 & 0.9 & \textbf{0.001} & \textbf{0.001} & 0.776 & 0.9 & 0.9 & 0.9 & \textbf{0.001} & \textbf{0.001} & \textbf{0.044} & 0.585 & \textbf{0.001} & \textbf{0.001} & 0.9 & 0.9 & 0.9 & \textbf{0.048} & 0.9 & 0.688 & 0.065 & 1.0 & 0.794 & 0.115 & 0.9 & 0.776 & 0.9 & 0.9 & 0.123 \\
ECOD & 0.9 & \textbf{0.001} & \textbf{0.019} & 0.9 & 0.241 & \textbf{0.001} & \textbf{0.002} & 0.9 & 0.9 & 0.121 & 0.9 & 0.9 & 0.085 & 0.9 & 0.9 & 0.169 & \textbf{0.009} & 0.9 & 0.9 & 0.9 & 0.9 & 0.241 & 0.9 & 0.9 & 0.794 & 1.0 & 0.9 & 0.9 & 0.9 & 0.633 & 0.9 & 0.9 \\
LMDD & 0.9 & \textbf{0.048} & 0.433 & 0.571 & \textbf{0.007} & \textbf{0.004} & 0.123 & 0.9 & 0.474 & \textbf{0.002} & 0.695 & 0.9 & 0.725 & 0.9 & 0.9 & 0.886 & 0.279 & 0.9 & 0.9 & 0.22 & 0.9 & \textbf{0.007} & 0.9 & 0.9 & 0.115 & 0.9 & 1.0 & 0.527 & 0.9 & 0.054 & 0.9 & 0.9 \\
gen2out & 0.9 & \textbf{0.001} & \textbf{0.001} & 0.9 & 0.9 & \textbf{0.001} & \textbf{0.001} & 0.9 & 0.9 & 0.9 & 0.9 & \textbf{0.017} & \textbf{0.001} & 0.304 & 0.9 & \textbf{0.001} & \textbf{0.001} & 0.9 & 0.9 & 0.9 & 0.322 & 0.9 & 0.9 & 0.389 & 0.9 & 0.9 & 0.527 & 1.0 & 0.9 & 0.9 & 0.9 & 0.541 \\
LUNAR & 0.9 & \textbf{0.001} & \textbf{0.021} & 0.9 & 0.226 & \textbf{0.001} & \textbf{0.002} & 0.9 & 0.9 & 0.112 & 0.9 & 0.9 & 0.091 & 0.9 & 0.9 & 0.181 & \textbf{0.01} & 0.9 & 0.9 & 0.9 & 0.9 & 0.226 & 0.9 & 0.9 & 0.776 & 0.9 & 0.9 & 0.9 & 1.0 & 0.615 & 0.9 & 0.9 \\
KDE & 0.666 & \textbf{0.001} & \textbf{0.001} & 0.9 & 0.9 & \textbf{0.001} & \textbf{0.001} & 0.615 & 0.9 & 0.9 & 0.9 & \textbf{0.001} & \textbf{0.001} & \textbf{0.018} & 0.416 & \textbf{0.001} & \textbf{0.001} & 0.9 & 0.9 & 0.9 & \textbf{0.02} & 0.9 & 0.527 & \textbf{0.028} & 0.9 & 0.633 & 0.054 & 0.9 & 0.615 & 1.0 & 0.9 & 0.058 \\
HBOS & 0.9 & \textbf{0.001} & \textbf{0.001} & 0.9 & 0.857 & \textbf{0.001} & \textbf{0.001} & 0.9 & 0.9 & 0.681 & 0.9 & 0.293 & \textbf{0.003} & 0.9 & 0.9 & \textbf{0.008} & \textbf{0.001} & 0.9 & 0.9 & 0.9 & 0.9 & 0.857 & 0.9 & 0.9 & 0.9 & 0.9 & 0.9 & 0.9 & 0.9 & 0.9 & 1.0 & 0.9 \\
SOD & 0.9 & \textbf{0.045} & 0.416 & 0.585 & \textbf{0.007} & \textbf{0.003} & 0.115 & 0.9 & 0.49 & \textbf{0.002} & 0.71 & 0.9 & 0.71 & 0.9 & 0.9 & 0.871 & 0.265 & 0.9 & 0.9 & 0.232 & 0.9 & \textbf{0.007} & 0.9 & 0.9 & 0.123 & 0.9 & 0.9 & 0.541 & 0.9 & 0.058 & 0.9 & 1.0 \\
\bottomrule
\end{tabular}

%% file: Tables/nemenyi_table_local.tex
\begin{tabular}{lllllllllllllllllllllllllllllllll}
\toprule
{} & {GMM} & {CBLOF} & {SO-GAAL} & {INNE} & {EIF} & {sb-DeepSVDD} & {ALAD} & {AE} & {MCD} & {kth-NN} & {COPOD} & {LOF} & {ODIN} & {ensemble-LOF} & {beta-VAE} & {COF} & {DeepSVDD} & {u-CBLOF} & {OCSVM} & {ABOD} & {PCA} & {kNN} & {VAE} & {LODA} & {IF} & {ECOD} & {LMDD} & {gen2out} & {LUNAR} & {KDE} & {HBOS} & {SOD} \\
\midrule
GMM & 1.0 & 0.855 & \textbf{0.039} & 0.9 & 0.9 & 0.434 & 0.128 & 0.155 & 0.9 & 0.9 & 0.125 & 0.9 & 0.9 & 0.9 & \textbf{0.034} & 0.9 & 0.647 & 0.9 & 0.452 & 0.9 & \textbf{0.01} & 0.9 & 0.81 & 0.499 & 0.9 & \textbf{0.026} & \textbf{0.039} & 0.869 & 0.9 & 0.9 & 0.855 & 0.9 \\
CBLOF & 0.855 & 1.0 & 0.9 & 0.9 & 0.9 & 0.9 & 0.9 & 0.9 & 0.9 & 0.165 & 0.9 & 0.364 & 0.9 & 0.476 & 0.9 & 0.559 & 0.9 & 0.9 & 0.9 & \textbf{0.029} & 0.9 & \textbf{0.021} & 0.9 & 0.9 & 0.9 & 0.9 & 0.9 & 0.9 & 0.855 & 0.452 & 0.9 & 0.434 \\
SO-GAAL & \textbf{0.039} & 0.9 & 1.0 & 0.795 & 0.9 & 0.9 & 0.9 & 0.9 & 0.107 & \textbf{0.001} & 0.9 & \textbf{0.002} & 0.232 & \textbf{0.004} & 0.9 & \textbf{0.007} & 0.9 & 0.633 & 0.9 & \textbf{0.001} & 0.9 & \textbf{0.001} & 0.9 & 0.9 & 0.9 & 0.9 & 0.9 & 0.9 & \textbf{0.039} & \textbf{0.004} & 0.9 & \textbf{0.003} \\
INNE & 0.9 & 0.9 & 0.795 & 1.0 & 0.9 & 0.9 & 0.9 & 0.9 & 0.9 & 0.825 & 0.9 & 0.9 & 0.9 & 0.9 & 0.766 & 0.9 & 0.9 & 0.9 & 0.9 & 0.452 & 0.559 & 0.381 & 0.9 & 0.9 & 0.9 & 0.721 & 0.795 & 0.9 & 0.9 & 0.9 & 0.9 & 0.9 \\
EIF & 0.9 & 0.9 & 0.9 & 0.9 & 1.0 & 0.9 & 0.9 & 0.9 & 0.9 & 0.692 & 0.9 & 0.9 & 0.9 & 0.9 & 0.899 & 0.9 & 0.9 & 0.9 & 0.9 & 0.3 & 0.692 & 0.244 & 0.9 & 0.9 & 0.9 & 0.855 & 0.9 & 0.9 & 0.9 & 0.9 & 0.9 & 0.9 \\
sb-DeepSVDD & 0.434 & 0.9 & 0.9 & 0.9 & 0.9 & 1.0 & 0.9 & 0.9 & 0.647 & \textbf{0.022} & 0.9 & 0.071 & 0.84 & 0.11 & 0.9 & 0.155 & 0.9 & 0.9 & 0.9 & \textbf{0.002} & 0.9 & \textbf{0.002} & 0.9 & 0.9 & 0.9 & 0.9 & 0.9 & 0.9 & 0.434 & 0.099 & 0.9 & 0.093 \\
ALAD & 0.128 & 0.9 & 0.9 & 0.9 & 0.9 & 0.9 & 1.0 & 0.9 & 0.285 & \textbf{0.003} & 0.9 & \textbf{0.011} & 0.499 & \textbf{0.02} & 0.9 & \textbf{0.031} & 0.9 & 0.884 & 0.9 & \textbf{0.001} & 0.9 & \textbf{0.001} & 0.9 & 0.9 & 0.9 & 0.9 & 0.9 & 0.9 & 0.128 & \textbf{0.017} & 0.9 & \textbf{0.016} \\
AE & 0.155 & 0.9 & 0.9 & 0.9 & 0.9 & 0.9 & 0.9 & 1.0 & 0.33 & \textbf{0.004} & 0.9 & \textbf{0.015} & 0.544 & \textbf{0.025} & 0.9 & \textbf{0.039} & 0.9 & 0.9 & 0.9 & \textbf{0.001} & 0.9 & \textbf{0.001} & 0.9 & 0.9 & 0.9 & 0.9 & 0.9 & 0.9 & 0.155 & \textbf{0.022} & 0.9 & \textbf{0.021} \\
MCD & 0.9 & 0.9 & 0.107 & 0.9 & 0.9 & 0.647 & 0.285 & 0.33 & 1.0 & 0.9 & 0.278 & 0.9 & 0.9 & 0.9 & 0.093 & 0.9 & 0.855 & 0.9 & 0.662 & 0.9 & \textbf{0.034} & 0.9 & 0.9 & 0.707 & 0.9 & 0.076 & 0.107 & 0.9 & 0.9 & 0.9 & 0.9 & 0.9 \\
kth-NN & 0.9 & 0.165 & \textbf{0.001} & 0.825 & 0.692 & \textbf{0.022} & \textbf{0.003} & \textbf{0.004} & 0.9 & 1.0 & \textbf{0.003} & 0.9 & 0.9 & 0.9 & \textbf{0.001} & 0.9 & 0.066 & 0.9 & \textbf{0.024} & 0.9 & \textbf{0.001} & 0.9 & 0.136 & \textbf{0.031} & 0.588 & \textbf{0.001} & \textbf{0.001} & 0.175 & 0.9 & 0.9 & 0.165 & 0.9 \\
COPOD & 0.125 & 0.9 & 0.9 & 0.9 & 0.9 & 0.9 & 0.9 & 0.9 & 0.278 & \textbf{0.003} & 1.0 & \textbf{0.011} & 0.492 & \textbf{0.019} & 0.9 & \textbf{0.03} & 0.9 & 0.877 & 0.9 & \textbf{0.001} & 0.9 & \textbf{0.001} & 0.9 & 0.9 & 0.9 & 0.9 & 0.9 & 0.9 & 0.125 & \textbf{0.017} & 0.9 & \textbf{0.015} \\
LOF & 0.9 & 0.364 & \textbf{0.002} & 0.9 & 0.9 & 0.071 & \textbf{0.011} & \textbf{0.015} & 0.9 & 0.9 & \textbf{0.011} & 1.0 & 0.9 & 0.9 & \textbf{0.002} & 0.9 & 0.175 & 0.9 & 0.076 & 0.9 & \textbf{0.001} & 0.9 & 0.314 & 0.093 & 0.81 & \textbf{0.001} & \textbf{0.002} & 0.381 & 0.9 & 0.9 & 0.364 & 0.9 \\
ODIN & 0.9 & 0.9 & 0.232 & 0.9 & 0.9 & 0.84 & 0.499 & 0.544 & 0.9 & 0.9 & 0.492 & 0.9 & 1.0 & 0.9 & 0.208 & 0.9 & 0.9 & 0.9 & 0.855 & 0.9 & 0.087 & 0.9 & 0.9 & 0.899 & 0.9 & 0.175 & 0.232 & 0.9 & 0.9 & 0.9 & 0.9 & 0.9 \\
ensemble-LOF & 0.9 & 0.476 & \textbf{0.004} & 0.9 & 0.9 & 0.11 & \textbf{0.02} & \textbf{0.025} & 0.9 & 0.9 & \textbf{0.019} & 0.9 & 0.9 & 1.0 & \textbf{0.004} & 0.9 & 0.251 & 0.9 & 0.118 & 0.9 & \textbf{0.001} & 0.9 & 0.426 & 0.141 & 0.9 & \textbf{0.003} & \textbf{0.004} & 0.492 & 0.9 & 0.9 & 0.476 & 0.9 \\
beta-VAE & \textbf{0.034} & 0.9 & 0.9 & 0.766 & 0.899 & 0.9 & 0.9 & 0.9 & 0.093 & \textbf{0.001} & 0.9 & \textbf{0.002} & 0.208 & \textbf{0.004} & 1.0 & \textbf{0.006} & 0.9 & 0.603 & 0.9 & \textbf{0.001} & 0.9 & \textbf{0.001} & 0.9 & 0.9 & 0.9 & 0.9 & 0.9 & 0.9 & \textbf{0.034} & \textbf{0.003} & 0.9 & \textbf{0.003} \\
COF & 0.9 & 0.559 & \textbf{0.007} & 0.9 & 0.9 & 0.155 & \textbf{0.031} & \textbf{0.039} & 0.9 & 0.9 & \textbf{0.03} & 0.9 & 0.9 & 0.9 & \textbf{0.006} & 1.0 & 0.33 & 0.9 & 0.165 & 0.9 & \textbf{0.002} & 0.9 & 0.514 & 0.196 & 0.9 & \textbf{0.005} & \textbf{0.007} & 0.573 & 0.9 & 0.9 & 0.559 & 0.9 \\
DeepSVDD & 0.647 & 0.9 & 0.9 & 0.9 & 0.9 & 0.9 & 0.9 & 0.9 & 0.855 & 0.066 & 0.9 & 0.175 & 0.9 & 0.251 & 0.9 & 0.33 & 1.0 & 0.9 & 0.9 & \textbf{0.009} & 0.9 & \textbf{0.006} & 0.9 & 0.9 & 0.9 & 0.9 & 0.9 & 0.9 & 0.647 & 0.232 & 0.9 & 0.219 \\
u-CBLOF & 0.9 & 0.9 & 0.633 & 0.9 & 0.9 & 0.9 & 0.884 & 0.9 & 0.9 & 0.9 & 0.877 & 0.9 & 0.9 & 0.9 & 0.603 & 0.9 & 0.9 & 1.0 & 0.9 & 0.618 & 0.381 & 0.559 & 0.9 & 0.9 & 0.9 & 0.559 & 0.633 & 0.9 & 0.9 & 0.9 & 0.9 & 0.9 \\
OCSVM & 0.452 & 0.9 & 0.9 & 0.9 & 0.9 & 0.9 & 0.9 & 0.9 & 0.662 & \textbf{0.024} & 0.9 & 0.076 & 0.855 & 0.118 & 0.9 & 0.165 & 0.9 & 0.9 & 1.0 & \textbf{0.003} & 0.9 & \textbf{0.002} & 0.9 & 0.9 & 0.9 & 0.9 & 0.9 & 0.9 & 0.452 & 0.107 & 0.9 & 0.099 \\
ABOD & 0.9 & \textbf{0.029} & \textbf{0.001} & 0.452 & 0.3 & \textbf{0.002} & \textbf{0.001} & \textbf{0.001} & 0.9 & 0.9 & \textbf{0.001} & 0.9 & 0.9 & 0.9 & \textbf{0.001} & 0.9 & \textbf{0.009} & 0.618 & \textbf{0.003} & 1.0 & \textbf{0.001} & 0.9 & \textbf{0.022} & \textbf{0.003} & 0.208 & \textbf{0.001} & \textbf{0.001} & \textbf{0.031} & 0.9 & 0.9 & \textbf{0.029} & 0.9 \\
PCA & \textbf{0.01} & 0.9 & 0.9 & 0.559 & 0.692 & 0.9 & 0.9 & 0.9 & \textbf{0.034} & \textbf{0.001} & 0.9 & \textbf{0.001} & 0.087 & \textbf{0.001} & 0.9 & \textbf{0.002} & 0.9 & 0.381 & 0.9 & \textbf{0.001} & 1.0 & \textbf{0.001} & 0.9 & 0.9 & 0.795 & 0.9 & 0.9 & 0.9 & \textbf{0.01} & \textbf{0.001} & 0.9 & \textbf{0.001} \\
kNN & 0.9 & \textbf{0.021} & \textbf{0.001} & 0.381 & 0.244 & \textbf{0.002} & \textbf{0.001} & \textbf{0.001} & 0.9 & 0.9 & \textbf{0.001} & 0.9 & 0.9 & 0.9 & \textbf{0.001} & 0.9 & \textbf{0.006} & 0.559 & \textbf{0.002} & 0.9 & \textbf{0.001} & 1.0 & \textbf{0.016} & \textbf{0.002} & 0.165 & \textbf{0.001} & \textbf{0.001} & \textbf{0.022} & 0.9 & 0.9 & \textbf{0.021} & 0.9 \\
VAE & 0.81 & 0.9 & 0.9 & 0.9 & 0.9 & 0.9 & 0.9 & 0.9 & 0.9 & 0.136 & 0.9 & 0.314 & 0.9 & 0.426 & 0.9 & 0.514 & 0.9 & 0.9 & 0.9 & \textbf{0.022} & 0.9 & \textbf{0.016} & 1.0 & 0.9 & 0.9 & 0.9 & 0.9 & 0.9 & 0.81 & 0.399 & 0.9 & 0.381 \\
LODA & 0.499 & 0.9 & 0.9 & 0.9 & 0.9 & 0.9 & 0.9 & 0.9 & 0.707 & \textbf{0.031} & 0.9 & 0.093 & 0.899 & 0.141 & 0.9 & 0.196 & 0.9 & 0.9 & 0.9 & \textbf{0.003} & 0.9 & \textbf{0.002} & 0.9 & 1.0 & 0.9 & 0.9 & 0.9 & 0.9 & 0.499 & 0.128 & 0.9 & 0.122 \\
IF & 0.9 & 0.9 & 0.9 & 0.9 & 0.9 & 0.9 & 0.9 & 0.9 & 0.9 & 0.588 & 0.9 & 0.81 & 0.9 & 0.9 & 0.9 & 0.9 & 0.9 & 0.9 & 0.9 & 0.208 & 0.795 & 0.165 & 0.9 & 0.9 & 1.0 & 0.9 & 0.9 & 0.9 & 0.9 & 0.884 & 0.9 & 0.869 \\
ECOD & \textbf{0.026} & 0.9 & 0.9 & 0.721 & 0.855 & 0.9 & 0.9 & 0.9 & 0.076 & \textbf{0.001} & 0.9 & \textbf{0.001} & 0.175 & \textbf{0.003} & 0.9 & \textbf{0.005} & 0.9 & 0.559 & 0.9 & \textbf{0.001} & 0.9 & \textbf{0.001} & 0.9 & 0.9 & 0.9 & 1.0 & 0.9 & 0.9 & \textbf{0.026} & \textbf{0.002} & 0.9 & \textbf{0.002} \\
LMDD & \textbf{0.039} & 0.9 & 0.9 & 0.795 & 0.9 & 0.9 & 0.9 & 0.9 & 0.107 & \textbf{0.001} & 0.9 & \textbf{0.002} & 0.232 & \textbf{0.004} & 0.9 & \textbf{0.007} & 0.9 & 0.633 & 0.9 & \textbf{0.001} & 0.9 & \textbf{0.001} & 0.9 & 0.9 & 0.9 & 0.9 & 1.0 & 0.9 & \textbf{0.039} & \textbf{0.004} & 0.9 & \textbf{0.003} \\
gen2out & 0.869 & 0.9 & 0.9 & 0.9 & 0.9 & 0.9 & 0.9 & 0.9 & 0.9 & 0.175 & 0.9 & 0.381 & 0.9 & 0.492 & 0.9 & 0.573 & 0.9 & 0.9 & 0.9 & \textbf{0.031} & 0.9 & \textbf{0.022} & 0.9 & 0.9 & 0.9 & 0.9 & 0.9 & 1.0 & 0.869 & 0.468 & 0.9 & 0.452 \\
LUNAR & 0.9 & 0.855 & \textbf{0.039} & 0.9 & 0.9 & 0.434 & 0.128 & 0.155 & 0.9 & 0.9 & 0.125 & 0.9 & 0.9 & 0.9 & \textbf{0.034} & 0.9 & 0.647 & 0.9 & 0.452 & 0.9 & \textbf{0.01} & 0.9 & 0.81 & 0.499 & 0.9 & \textbf{0.026} & \textbf{0.039} & 0.869 & 1.0 & 0.9 & 0.855 & 0.9 \\
KDE & 0.9 & 0.452 & \textbf{0.004} & 0.9 & 0.9 & 0.099 & \textbf{0.017} & \textbf{0.022} & 0.9 & 0.9 & \textbf{0.017} & 0.9 & 0.9 & 0.9 & \textbf{0.003} & 0.9 & 0.232 & 0.9 & 0.107 & 0.9 & \textbf{0.001} & 0.9 & 0.399 & 0.128 & 0.884 & \textbf{0.002} & \textbf{0.004} & 0.468 & 0.9 & 1.0 & 0.452 & 0.9 \\
HBOS & 0.855 & 0.9 & 0.9 & 0.9 & 0.9 & 0.9 & 0.9 & 0.9 & 0.9 & 0.165 & 0.9 & 0.364 & 0.9 & 0.476 & 0.9 & 0.559 & 0.9 & 0.9 & 0.9 & \textbf{0.029} & 0.9 & \textbf{0.021} & 0.9 & 0.9 & 0.9 & 0.9 & 0.9 & 0.9 & 0.855 & 0.452 & 1.0 & 0.434 \\
SOD & 0.9 & 0.434 & \textbf{0.003} & 0.9 & 0.9 & 0.093 & \textbf{0.016} & \textbf{0.021} & 0.9 & 0.9 & \textbf{0.015} & 0.9 & 0.9 & 0.9 & \textbf{0.003} & 0.9 & 0.219 & 0.9 & 0.099 & 0.9 & \textbf{0.001} & 0.9 & 0.381 & 0.122 & 0.869 & \textbf{0.002} & \textbf{0.003} & 0.452 & 0.9 & 0.9 & 0.434 & 1.0 \\
\bottomrule
\end{tabular}

%% file: Tables/nemenyi_table_global.tex
\begin{tabular}{lllllllllllllllllllllllllllllllll}
\toprule
{} & {GMM} & {CBLOF} & {SO-GAAL} & {INNE} & {EIF} & {sb-DeepSVDD} & {ALAD} & {AE} & {MCD} & {kth-NN} & {COPOD} & {LOF} & {ODIN} & {ensemble-LOF} & {beta-VAE} & {COF} & {DeepSVDD} & {u-CBLOF} & {OCSVM} & {ABOD} & {PCA} & {kNN} & {VAE} & {LODA} & {IF} & {ECOD} & {LMDD} & {gen2out} & {LUNAR} & {KDE} & {HBOS} & {SOD} \\
\midrule
GMM & 1.0 & \textbf{0.007} & 0.757 & 0.613 & \textbf{0.002} & \textbf{0.001} & 0.2 & 0.9 & 0.9 & 0.165 & 0.063 & 0.35 & \textbf{0.034} & 0.9 & 0.9 & \textbf{0.01} & 0.154 & 0.9 & 0.791 & 0.9 & 0.9 & 0.605 & 0.9 & 0.9 & 0.052 & 0.9 & 0.9 & 0.187 & 0.9 & 0.622 & 0.9 & 0.9 \\
CBLOF & \textbf{0.007} & 1.0 & 0.9 & \textbf{0.001} & \textbf{0.001} & 0.9 & 0.9 & \textbf{0.001} & \textbf{0.001} & \textbf{0.001} & \textbf{0.001} & 0.9 & 0.9 & 0.866 & \textbf{0.001} & 0.9 & 0.9 & \textbf{0.001} & \textbf{0.001} & \textbf{0.001} & \textbf{0.001} & \textbf{0.001} & \textbf{0.001} & \textbf{0.004} & \textbf{0.001} & \textbf{0.001} & \textbf{0.001} & \textbf{0.001} & \textbf{0.01} & \textbf{0.001} & \textbf{0.001} & 0.639 \\
SO-GAAL & 0.757 & 0.9 & 1.0 & \textbf{0.001} & \textbf{0.001} & 0.9 & 0.9 & \textbf{0.005} & \textbf{0.001} & \textbf{0.001} & \textbf{0.001} & 0.9 & 0.9 & 0.9 & \textbf{0.006} & 0.9 & 0.9 & \textbf{0.007} & \textbf{0.001} & \textbf{0.014} & 0.2 & \textbf{0.001} & 0.073 & 0.647 & \textbf{0.001} & \textbf{0.001} & 0.148 & \textbf{0.001} & 0.816 & \textbf{0.001} & \textbf{0.001} & 0.9 \\
INNE & 0.613 & \textbf{0.001} & \textbf{0.001} & 1.0 & 0.9 & \textbf{0.001} & \textbf{0.001} & 0.9 & 0.9 & 0.9 & 0.9 & \textbf{0.001} & \textbf{0.001} & \textbf{0.001} & 0.9 & \textbf{0.001} & \textbf{0.001} & 0.9 & 0.9 & 0.9 & 0.9 & 0.9 & 0.9 & 0.723 & 0.9 & 0.9 & 0.9 & 0.9 & 0.554 & 0.9 & 0.9 & \textbf{0.006} \\
EIF & \textbf{0.002} & \textbf{0.001} & \textbf{0.001} & 0.9 & 1.0 & \textbf{0.001} & \textbf{0.001} & 0.613 & 0.892 & 0.9 & 0.9 & \textbf{0.001} & \textbf{0.001} & \textbf{0.001} & 0.58 & \textbf{0.001} & \textbf{0.001} & 0.554 & 0.9 & 0.44 & 0.05 & 0.9 & 0.148 & \textbf{0.004} & 0.9 & 0.845 & 0.073 & 0.9 & \textbf{0.001} & 0.9 & 0.9 & \textbf{0.001} \\
sb-DeepSVDD & \textbf{0.001} & 0.9 & 0.9 & \textbf{0.001} & \textbf{0.001} & 1.0 & 0.9 & \textbf{0.001} & \textbf{0.001} & \textbf{0.001} & \textbf{0.001} & 0.9 & 0.9 & 0.597 & \textbf{0.001} & 0.9 & 0.9 & \textbf{0.001} & \textbf{0.001} & \textbf{0.001} & \textbf{0.001} & \textbf{0.001} & \textbf{0.001} & \textbf{0.001} & \textbf{0.001} & \textbf{0.001} & \textbf{0.001} & \textbf{0.001} & \textbf{0.002} & \textbf{0.001} & \textbf{0.001} & 0.35 \\
ALAD & 0.2 & 0.9 & 0.9 & \textbf{0.001} & \textbf{0.001} & 0.9 & 1.0 & \textbf{0.001} & \textbf{0.001} & \textbf{0.001} & \textbf{0.001} & 0.9 & 0.9 & 0.9 & \textbf{0.001} & 0.9 & 0.9 & \textbf{0.001} & \textbf{0.001} & \textbf{0.001} & \textbf{0.014} & \textbf{0.001} & \textbf{0.003} & 0.128 & \textbf{0.001} & \textbf{0.001} & \textbf{0.009} & \textbf{0.001} & 0.249 & \textbf{0.001} & \textbf{0.001} & 0.9 \\
AE & 0.9 & \textbf{0.001} & \textbf{0.005} & 0.9 & 0.613 & \textbf{0.001} & \textbf{0.001} & 1.0 & 0.9 & 0.9 & 0.9 & \textbf{0.001} & \textbf{0.001} & \textbf{0.048} & 0.9 & \textbf{0.001} & \textbf{0.001} & 0.9 & 0.9 & 0.9 & 0.9 & 0.9 & 0.9 & 0.9 & 0.9 & 0.9 & 0.9 & 0.9 & 0.9 & 0.9 & 0.9 & 0.137 \\
MCD & 0.9 & \textbf{0.001} & \textbf{0.001} & 0.9 & 0.892 & \textbf{0.001} & \textbf{0.001} & 0.9 & 1.0 & 0.9 & 0.9 & \textbf{0.001} & \textbf{0.001} & \textbf{0.01} & 0.9 & \textbf{0.001} & \textbf{0.001} & 0.9 & 0.9 & 0.9 & 0.9 & 0.9 & 0.9 & 0.9 & 0.9 & 0.9 & 0.9 & 0.9 & 0.866 & 0.9 & 0.9 & \textbf{0.037} \\
kth-NN & 0.165 & \textbf{0.001} & \textbf{0.001} & 0.9 & 0.9 & \textbf{0.001} & \textbf{0.001} & 0.9 & 0.9 & 1.0 & 0.9 & \textbf{0.001} & \textbf{0.001} & \textbf{0.001} & 0.9 & \textbf{0.001} & \textbf{0.001} & 0.9 & 0.9 & 0.9 & 0.706 & 0.9 & 0.9 & 0.249 & 0.9 & 0.9 & 0.782 & 0.9 & 0.128 & 0.9 & 0.9 & \textbf{0.001} \\
COPOD & 0.063 & \textbf{0.001} & \textbf{0.001} & 0.9 & 0.9 & \textbf{0.001} & \textbf{0.001} & 0.9 & 0.9 & 0.9 & 1.0 & \textbf{0.001} & \textbf{0.001} & \textbf{0.001} & 0.9 & \textbf{0.001} & \textbf{0.001} & 0.9 & 0.9 & 0.9 & 0.491 & 0.9 & 0.727 & 0.105 & 0.9 & 0.9 & 0.567 & 0.9 & \textbf{0.047} & 0.9 & 0.9 & \textbf{0.001} \\
LOF & 0.35 & 0.9 & 0.9 & \textbf{0.001} & \textbf{0.001} & 0.9 & 0.9 & \textbf{0.001} & \textbf{0.001} & \textbf{0.001} & \textbf{0.001} & 1.0 & 0.9 & 0.9 & \textbf{0.001} & 0.9 & 0.9 & \textbf{0.001} & \textbf{0.001} & \textbf{0.001} & \textbf{0.034} & \textbf{0.001} & \textbf{0.009} & 0.241 & \textbf{0.001} & \textbf{0.001} & \textbf{0.022} & \textbf{0.001} & 0.42 & \textbf{0.001} & \textbf{0.001} & 0.9 \\
ODIN & \textbf{0.034} & 0.9 & 0.9 & \textbf{0.001} & \textbf{0.001} & 0.9 & 0.9 & \textbf{0.001} & \textbf{0.001} & \textbf{0.001} & \textbf{0.001} & 0.9 & 1.0 & 0.9 & \textbf{0.001} & 0.9 & 0.9 & \textbf{0.001} & \textbf{0.001} & \textbf{0.001} & \textbf{0.001} & \textbf{0.001} & \textbf{0.001} & \textbf{0.019} & \textbf{0.001} & \textbf{0.001} & \textbf{0.001} & \textbf{0.001} & \textbf{0.046} & \textbf{0.001} & \textbf{0.001} & 0.9 \\
ensemble-LOF & 0.9 & 0.866 & 0.9 & \textbf{0.001} & \textbf{0.001} & 0.597 & 0.9 & \textbf{0.048} & \textbf{0.01} & \textbf{0.001} & \textbf{0.001} & 0.9 & 0.9 & 1.0 & 0.057 & 0.9 & 0.9 & 0.065 & \textbf{0.004} & 0.107 & 0.605 & \textbf{0.001} & 0.35 & 0.9 & \textbf{0.001} & \textbf{0.014} & 0.529 & \textbf{0.001} & 0.9 & \textbf{0.001} & \textbf{0.009} & 0.9 \\
beta-VAE & 0.9 & \textbf{0.001} & \textbf{0.006} & 0.9 & 0.58 & \textbf{0.001} & \textbf{0.001} & 0.9 & 0.9 & 0.9 & 0.9 & \textbf{0.001} & \textbf{0.001} & 0.057 & 1.0 & \textbf{0.001} & \textbf{0.001} & 0.9 & 0.9 & 0.9 & 0.9 & 0.9 & 0.9 & 0.9 & 0.9 & 0.9 & 0.9 & 0.9 & 0.9 & 0.9 & 0.9 & 0.159 \\
COF & \textbf{0.01} & 0.9 & 0.9 & \textbf{0.001} & \textbf{0.001} & 0.9 & 0.9 & \textbf{0.001} & \textbf{0.001} & \textbf{0.001} & \textbf{0.001} & 0.9 & 0.9 & 0.9 & \textbf{0.001} & 1.0 & 0.9 & \textbf{0.001} & \textbf{0.001} & \textbf{0.001} & \textbf{0.001} & \textbf{0.001} & \textbf{0.001} & \textbf{0.005} & \textbf{0.001} & \textbf{0.001} & \textbf{0.001} & \textbf{0.001} & \textbf{0.015} & \textbf{0.001} & \textbf{0.001} & 0.698 \\
DeepSVDD & 0.154 & 0.9 & 0.9 & \textbf{0.001} & \textbf{0.001} & 0.9 & 0.9 & \textbf{0.001} & \textbf{0.001} & \textbf{0.001} & \textbf{0.001} & 0.9 & 0.9 & 0.9 & \textbf{0.001} & 0.9 & 1.0 & \textbf{0.001} & \textbf{0.001} & \textbf{0.001} & \textbf{0.009} & \textbf{0.001} & \textbf{0.002} & 0.095 & \textbf{0.001} & \textbf{0.001} & \textbf{0.006} & \textbf{0.001} & 0.194 & \textbf{0.001} & \textbf{0.001} & 0.9 \\
u-CBLOF & 0.9 & \textbf{0.001} & \textbf{0.007} & 0.9 & 0.554 & \textbf{0.001} & \textbf{0.001} & 0.9 & 0.9 & 0.9 & 0.9 & \textbf{0.001} & \textbf{0.001} & 0.065 & 0.9 & \textbf{0.001} & \textbf{0.001} & 1.0 & 0.9 & 0.9 & 0.9 & 0.9 & 0.9 & 0.9 & 0.9 & 0.9 & 0.9 & 0.9 & 0.9 & 0.9 & 0.9 & 0.175 \\
OCSVM & 0.791 & \textbf{0.001} & \textbf{0.001} & 0.9 & 0.9 & \textbf{0.001} & \textbf{0.001} & 0.9 & 0.9 & 0.9 & 0.9 & \textbf{0.001} & \textbf{0.001} & \textbf{0.004} & 0.9 & \textbf{0.001} & \textbf{0.001} & 0.9 & 1.0 & 0.9 & 0.9 & 0.9 & 0.9 & 0.9 & 0.9 & 0.9 & 0.9 & 0.9 & 0.731 & 0.9 & 0.9 & \textbf{0.018} \\
ABOD & 0.9 & \textbf{0.001} & \textbf{0.014} & 0.9 & 0.44 & \textbf{0.001} & \textbf{0.001} & 0.9 & 0.9 & 0.9 & 0.9 & \textbf{0.001} & \textbf{0.001} & 0.107 & 0.9 & \textbf{0.001} & \textbf{0.001} & 0.9 & 0.9 & 1.0 & 0.9 & 0.9 & 0.9 & 0.9 & 0.9 & 0.9 & 0.9 & 0.9 & 0.9 & 0.9 & 0.9 & 0.264 \\
PCA & 0.9 & \textbf{0.001} & 0.2 & 0.9 & 0.05 & \textbf{0.001} & \textbf{0.014} & 0.9 & 0.9 & 0.706 & 0.491 & \textbf{0.034} & \textbf{0.001} & 0.605 & 0.9 & \textbf{0.001} & \textbf{0.009} & 0.9 & 0.9 & 0.9 & 1.0 & 0.9 & 0.9 & 0.9 & 0.45 & 0.9 & 0.9 & 0.74 & 0.9 & 0.9 & 0.9 & 0.833 \\
kNN & 0.605 & \textbf{0.001} & \textbf{0.001} & 0.9 & 0.9 & \textbf{0.001} & \textbf{0.001} & 0.9 & 0.9 & 0.9 & 0.9 & \textbf{0.001} & \textbf{0.001} & \textbf{0.001} & 0.9 & \textbf{0.001} & \textbf{0.001} & 0.9 & 0.9 & 0.9 & 0.9 & 1.0 & 0.9 & 0.715 & 0.9 & 0.9 & 0.9 & 0.9 & 0.546 & 0.9 & 0.9 & \textbf{0.006} \\
VAE & 0.9 & \textbf{0.001} & 0.073 & 0.9 & 0.148 & \textbf{0.001} & \textbf{0.003} & 0.9 & 0.9 & 0.9 & 0.727 & \textbf{0.009} & \textbf{0.001} & 0.35 & 0.9 & \textbf{0.001} & \textbf{0.002} & 0.9 & 0.9 & 0.9 & 0.9 & 0.9 & 1.0 & 0.9 & 0.689 & 0.9 & 0.9 & 0.9 & 0.9 & 0.9 & 0.9 & 0.597 \\
LODA & 0.9 & \textbf{0.004} & 0.647 & 0.723 & \textbf{0.004} & \textbf{0.001} & 0.128 & 0.9 & 0.9 & 0.249 & 0.105 & 0.241 & \textbf{0.019} & 0.9 & 0.9 & \textbf{0.005} & 0.095 & 0.9 & 0.9 & 0.9 & 0.9 & 0.715 & 0.9 & 1.0 & 0.088 & 0.9 & 0.9 & 0.28 & 0.9 & 0.731 & 0.9 & 0.9 \\
IF & 0.052 & \textbf{0.001} & \textbf{0.001} & 0.9 & 0.9 & \textbf{0.001} & \textbf{0.001} & 0.9 & 0.9 & 0.9 & 0.9 & \textbf{0.001} & \textbf{0.001} & \textbf{0.001} & 0.9 & \textbf{0.001} & \textbf{0.001} & 0.9 & 0.9 & 0.9 & 0.45 & 0.9 & 0.689 & 0.088 & 1.0 & 0.9 & 0.529 & 0.9 & \textbf{0.039} & 0.9 & 0.9 & \textbf{0.001} \\
ECOD & 0.9 & \textbf{0.001} & \textbf{0.001} & 0.9 & 0.845 & \textbf{0.001} & \textbf{0.001} & 0.9 & 0.9 & 0.9 & 0.9 & \textbf{0.001} & \textbf{0.001} & \textbf{0.014} & 0.9 & \textbf{0.001} & \textbf{0.001} & 0.9 & 0.9 & 0.9 & 0.9 & 0.9 & 0.9 & 0.9 & 0.9 & 1.0 & 0.9 & 0.9 & 0.9 & 0.9 & 0.9 & \textbf{0.047} \\
LMDD & 0.9 & \textbf{0.001} & 0.148 & 0.9 & 0.073 & \textbf{0.001} & \textbf{0.009} & 0.9 & 0.9 & 0.782 & 0.567 & \textbf{0.022} & \textbf{0.001} & 0.529 & 0.9 & \textbf{0.001} & \textbf{0.006} & 0.9 & 0.9 & 0.9 & 0.9 & 0.9 & 0.9 & 0.9 & 0.529 & 0.9 & 1.0 & 0.816 & 0.9 & 0.9 & 0.9 & 0.757 \\
gen2out & 0.187 & \textbf{0.001} & \textbf{0.001} & 0.9 & 0.9 & \textbf{0.001} & \textbf{0.001} & 0.9 & 0.9 & 0.9 & 0.9 & \textbf{0.001} & \textbf{0.001} & \textbf{0.001} & 0.9 & \textbf{0.001} & \textbf{0.001} & 0.9 & 0.9 & 0.9 & 0.74 & 0.9 & 0.9 & 0.28 & 0.9 & 0.9 & 0.816 & 1.0 & 0.148 & 0.9 & 0.9 & \textbf{0.001} \\
LUNAR & 0.9 & \textbf{0.01} & 0.816 & 0.554 & \textbf{0.001} & \textbf{0.002} & 0.249 & 0.9 & 0.866 & 0.128 & \textbf{0.047} & 0.42 & \textbf{0.046} & 0.9 & 0.9 & \textbf{0.015} & 0.194 & 0.9 & 0.731 & 0.9 & 0.9 & 0.546 & 0.9 & 0.9 & \textbf{0.039} & 0.9 & 0.9 & 0.148 & 1.0 & 0.563 & 0.841 & 0.9 \\
KDE & 0.622 & \textbf{0.001} & \textbf{0.001} & 0.9 & 0.9 & \textbf{0.001} & \textbf{0.001} & 0.9 & 0.9 & 0.9 & 0.9 & \textbf{0.001} & \textbf{0.001} & \textbf{0.001} & 0.9 & \textbf{0.001} & \textbf{0.001} & 0.9 & 0.9 & 0.9 & 0.9 & 0.9 & 0.9 & 0.731 & 0.9 & 0.9 & 0.9 & 0.9 & 0.563 & 1.0 & 0.9 & \textbf{0.007} \\
HBOS & 0.9 & \textbf{0.001} & \textbf{0.001} & 0.9 & 0.9 & \textbf{0.001} & \textbf{0.001} & 0.9 & 0.9 & 0.9 & 0.9 & \textbf{0.001} & \textbf{0.001} & \textbf{0.009} & 0.9 & \textbf{0.001} & \textbf{0.001} & 0.9 & 0.9 & 0.9 & 0.9 & 0.9 & 0.9 & 0.9 & 0.9 & 0.9 & 0.9 & 0.9 & 0.841 & 0.9 & 1.0 & \textbf{0.032} \\
SOD & 0.9 & 0.639 & 0.9 & \textbf{0.006} & \textbf{0.001} & 0.35 & 0.9 & 0.137 & \textbf{0.037} & \textbf{0.001} & \textbf{0.001} & 0.9 & 0.9 & 0.9 & 0.159 & 0.698 & 0.9 & 0.175 & \textbf{0.018} & 0.264 & 0.833 & \textbf{0.006} & 0.597 & 0.9 & \textbf{0.001} & \textbf{0.047} & 0.757 & \textbf{0.001} & 0.9 & \textbf{0.007} & \textbf{0.032} & 1.0 \\
\bottomrule
\end{tabular}